\documentclass{article}

\usepackage{arxiv}

\usepackage[utf8]{inputenc} % allow utf-8 input
\usepackage[T1]{fontenc}    % use 8-bit T1 fonts
\usepackage{hyperref}       % hyperlinks
\usepackage{url}            % simple URL typesetting
\usepackage{booktabs}       % professional-quality tables
\usepackage{amsfonts}       % blackboard math symbols
\usepackage{nicefrac}       % compact symbols for 1/2, etc.
\usepackage{microtype}      % microtypography
\usepackage{lipsum}		% Can be removed after putting your text content
\usepackage{graphicx}

\usepackage{doi}

\usepackage{lineno,hyperref}

\usepackage{fancyhdr}

\usepackage{amsmath,amssymb,amsfonts}
\usepackage{subfigure}
\usepackage{algorithmic}
\usepackage{graphicx}
\usepackage{textcomp}
\usepackage{xcolor}
\usepackage{multirow}
\usepackage[normalem]{ulem}
\usepackage{float}
\usepackage[export]{adjustbox}
\usepackage{booktabs} 
\modulolinenumbers[5]

\title{Automatic CT Segmentation from Bounding Box Annotations using Convolutional Neural Networks}

\author{Yuanpeng Liu\\
	Institute of Applied Mathematics and Image Processing\\
	Zhejiang University\\
	Hangzhou, Zhejiang, 310058, China \\
	\texttt{liuyp23@gmail.com} \\
	\And
	{Qinglei Hui} \\
	School of Mathematical Sciences\\
	Zhejiang University\\
	Hangzhou, Zhejiang, 310058, China \\
	\texttt{qlhui@zju.edu.cn} \\
	\AND
	Zhiyi Peng \\
	Department of Radiology \\
	The First Affiliated Hospital, Zhejiang University School of Medicine \\
	Hangzhou, Zhejiang, 310003, China\\
	 \texttt{1190020@zju.edu.cn} \\
	\And
	Shaolin Gong \\
	Department of Radiology \\
	The First Affiliated Hospital, Zhejiang University School of Medicine \\
	Hangzhou, Zhejiang, 310003, China\\
	\texttt{gongshaolin@foxmail.com} \\
	\And
	{Dexing Kong} \\
	School of Mathematical Sciences\\
	Zhejiang University\\
	Hangzhou, Zhejiang, 310058, China \\
	\texttt{dxkong@zju.edu.cn} \\
}

\renewcommand{\shorttitle}{\textit{arXiv} Template}

\hypersetup{
pdftitle={A template for the arxiv style},
pdfsubject={q-bio.NC, q-bio.QM},
pdfauthor={David S.~Hippocampus, Elias D.~Striatum},
pdfkeywords={First keyword, Second keyword, More},
}

\begin{document}
\maketitle

\begin{abstract}
	Accurate segmentation for medical images is important for clinical diagnosis. Existing automatic segmentation methods are mainly based on fully supervised learning and have an extremely high demand for precise annotations, which are very costly and time-consuming to obtain. To address this problem, we proposed an automatic computerized tomography segmentation method based on weakly supervised learning, by which one could train an accurate segmentation model only with weak annotations in the form of bounding boxes. The proposed method is composed of two steps: 1) generating pseudo masks with bounding box annotations by k-means clustering, and 2) iteratively training a 3D U-Net convolutional neural network as a segmentation model. Some data pre-processing methods are used to improve performance. The method was validated on four datasets containing three types of organs with a total of 627 CT volumes. For liver, spleen and kidney segmentation, it achieved an accuracy of 95.19\%, 92.11\%, and 91.45\%, respectively. Experimental results demonstrate that our method is accurate, efficient, and suitable for clinical use.
\end{abstract}

% keywords can be removed

\section{Introduction}
Automatic organ segmentation from computerized tomography (CT) is of great importance for disease diagnosis and computer-aided surgery. As one of the most common radiographic techniques, CT takes advantage of the differences in X-ray attenuation of different tissues in human bodies and reflects them as image intensity in tomographic images. However, due to the fuzzy boundaries between different organs and tissues, the variability of organ morphology, and the diversity of different individuals of the same organ or tissue, CT image segmentation has always been a challenging task. Furthermore, the appearance of lesions or tumors makes this task even more difficult. 

With the high-speed development of deep learning and the improvement of graphics processing units (GPU) performance in recent years, the accuracy of medical image segmentation has improved rapidly with the help of convolutional neural networks (CNN). The parameters in CNN are iteratively updated, so that extracted features gradually match the accurate ground-truth annotations as closely as possible, which is so-called fully supervised learning. As one of the most representative fully supervised segmentation methods based on CNN, Fully Convolutional Networks (FCN) proposed by Long et al. \cite{7298965} took advantage of the encoder-decoder structure to generate probability maps that contained prediction for each pixel. In 2015, U-net, proposed by Ronneberger et al.\cite{ronneberger2015u}, added concatenated skip connections to the encoder-decoder structure to fuse high and low-dimensional information to improve feature representation. U-net reached state-of-the-art for a long time, and its excellent performance makes it one of the most common backbone networks today. A year later, Ö Çiçek et al.\cite{10.1007/978-3-319-46723-8_49} proposed 3D-Unet, which uses 3D convolution to capture information between slices in 3D volumes that are important in medical images. In 2018, Ozan Oktayet al.\cite{oktay2018attention} added the attention mechanism\cite{mnih2014recurrent} to Unet to make the network focus on the target’s shape and size. Fabian Isensee et al.\cite{isensee2018nnu} proposed an adaptive U-net network using the idea of a two-stage cascade, which has achieved excellent results on medical images of different organs. Despite its excellent performance, fully supervised learning based segmentation methods excessively rely on pixel-wise manual annotations, which required experienced radiologists to delineate the targets on high-resolution medical images pixel by pixel, making it extremely inefficient and costly.

To solve this problem, many researchers have devoted more attention to weakly supervised learning. According to the different types of objectives and annotations, weak supervision can be categorized into three types: incomplete supervision, inexact supervision, and inaccurate supervision\cite{10.1093/nsr/nwx106}. Incomplete supervision refers to the situation in which there is only a small portion of training samples with annotations. In this situation, the common solutions are divided into active learning and semi-supervised learning according to the presence of external interventions. Active learning requires periodic external intervention to correct the learning direction of the model. Top et al.\cite{10.1007/978-3-642-23626-6_74} proposed a novel method for applying active learning strategies to interactive 3D image segmentation. Wang et al.\cite{WANG201934} proposed an interactive medical image segmentation model using deep learning. Semi-supervised learning does not rely on human intervention but on the distribution of the unlabeled data to make predictions. Li et al.\cite{li2018semi} used the idea proposed by Laine et al.\cite{laine2016temporal}to drive semi-supervised learning by utilizing the inconsistency of CNN outputs under different regularizations. Yu et al.\cite{10.1007/978-3-030-32245-8_67} applied the Mean-teacher model\cite{10.5555/3294771.3294885} to medical image segmentation, using uncertainty to guide the model to learn from unlabeled data. However, for incomplete supervised image segmentation, precisely annotated training data are still necessary.

Inexact supervision means that only coarse-grained labels are available for the training data. In this case, for image segmentation tasks, the given annotations are usually in the form of image-level tags. The most commonly used method in inexact supervised learning is multiple instance learning (MIL)\cite{DIETTERICH199731}. An image is regarded as a bag consisting of multiple instances, and by building a classifier, the label of each instance is predicted using the label of the bag. Cinbis et al.\cite{6909705} proposed a multi-fold MIL method to performs image segmentation with image-level tags. Pinheiro et al.\cite{7298780} combine multi-instance learning and convolutional neural networks for semantic segmentation. MIL has achieved good performance in image classification and object detection. Nevertheless, its accuracy in semantic segmentation tasks cannot match that of fully supervised learning.

The third type of weak supervision is called inaccurate supervision. In this scenario, the annotations in the training set are usually in the form of bounding boxes or scribbles, etc. They are not completely accurate and contain some false positive and false negative noise. Common solutions are refining the initial weak labels and learning with noise. In 2015, Dai et al.\cite{7410548} used generated proposal masks to update the labels and network parameters iteratively. In 2016, Rajchl et al.\cite{7739993} combined GrabCut\cite{10.1145/1015706.1015720} and conditional random fields (CRF)\cite{lafferty2001conditional} to drive the CNN training for segmenting Magnetic resonance imaging of brain and lung. Khoreva et al.\cite{8099664} train foreground segmentation network directly with mask generated by GrabCut and the weak supervision result reached ~95\% of the quality of the fully supervised model.

In this study, we proposed a precise organ segmentation method for CT images to handle the scenario of inaccurate supervision. After a series of machine learning and image processing operations on the initial bounding box annotations, we trained a neural network to obtain pixel-wise segmentation results. We conducted experiments and tests on four datasets containing three types of organs with a total of 627 CT volumes, and the results achieved a remarkable level of performance.

Compared to existing algorithms, there are two novelties in our work:
\begin{itemize}
	\item We proposed a novel approach to generate pseudo masks from bounding box annotations. After a series of preprocessing operations such as k-means clustering, morphological operations, and removal of connected components, the pseudo masks are generated as the initial labels to train the convolutional neural network.
	\item We designed a new CNN segmentation model called BA-Unet. While taking advantage of the encoder-decoder structure, bottleneck blocks and attention mechanism, the network additionally incorporates features from other dimensional encoders during skip connections when decoding high-dimensional features.
	\item The results on four clinical datasets show that our method achieves good performance in three different organ segmentation tasks and demonstrates good cross-domain performance through external validation.
\end{itemize}
This paper is organized in the following manner. Section \ref{sec-2} introduces the datasets used in this study. Section \ref{sec-3} states the methodology in detail, including the generation of bounding box annotations, pseudo masks, and Convolutional segmentation neural network details. Section \ref{sec-4} illustrates the experimental setup while Section \ref{sec-5} presents and analyzes the experimental results. Finally, We discuss the experimental results in Section \ref{sec-6} and summarize the entire paper in Section \ref{sec-7}.

\section{Materials}
\label{sec-2}
In this study, we collected a total of 627 CT images of livers, spleens, and kidneys. The liver dataset contains 245 CT volumes collected from the First Affiliated Hospital, Zhejiang University School of Medicine. The dataset includes cases in the unenhanced phase, arterial phase, portal phase, and delayed phase. Each CT scan has the same axial dimensions of 512 $\times$ 512 and slice number of 40. The pixel spacing ranges from 6.8 to 9.8 mm in the x-y direction, and the slice thickness is 5mm constantly. Each case in this dataset is manually annotated by an experienced radiologist. The spleen dataset is from Medical Segmentation Decathlon (MSD)\cite{simpson2019large} Challenge. Forty-one portal venous phase CT scans are included with annotations and have the same axial dimensions of 512 $\times$ 512 with slice numbers ranging from 31 to 168. The pixel spacing ranges from 6.1 to 9.8 mm in the x–y direction, and the slice distance ranges from 1.5 to 8 mm. The kidneys dataset is from 2019 Kidney Tumor Segmentation (KiTS)\cite{heller2019kits19} Challenge and 210 abdominal CT images in the late-arterial phase are available. All scans have the same axial dimensions of 512 $\times$ 512 with slice numbers ranging from 29 to 1059. The pixel spacing ranges from 4.4 to 10.0 mm in the x–y direction, and the slice thickness ranges from 0.5 to 5 mm. In addition, 131 liver CT scans from Liver Tumor Segmentation (LiTS)\cite{bilic2019liver} Challenge are used as external validation data. The image resolution ranges from 0.56 mm to 1.0 mm in axial and 0.45 mm to 6.0mm in the z direction, with slice numbers ranging from 42 to 1026. Most images in the study are pathologic and include tumors, lesions, or cysts of different sizes.

\section{Methods}
\label{sec-3}
\subsection{Pseudo masks generation}

The first step of our algorithm is to generate pseudo masks of the region of interests (ROI) which contain the target organs by k-means clustering\cite{https://doi.org/10.2307/2346830} in color space. Given a CT slice, k-means clustering partitions all the pixels into $k$ sets $\textbf{S}=\{S_1,  …, S_k\}$ such that the within-cluster sum of squares (WCSS) is minimized. The objective reads
\begin{equation}
	\arg \min_{\textbf{S}} \sum_{i=1}^{k}\sum_{\textbf{x}\in \textit{S}_{i}} {\lVert \textbf{x} - \boldsymbol{\mu} \rVert}^{2}
\end{equation}
Where $\boldsymbol{\mu}$ is the mean of pixel values in $S_i$. Since CT values of different organs and tissues are not identical, when the window width and level are well selected, the k-means algorithm can roughly separate organs and tissues in one image into k classes.

\begin{figure*}[h]
	\centerline{\includegraphics[width=6.8 in]{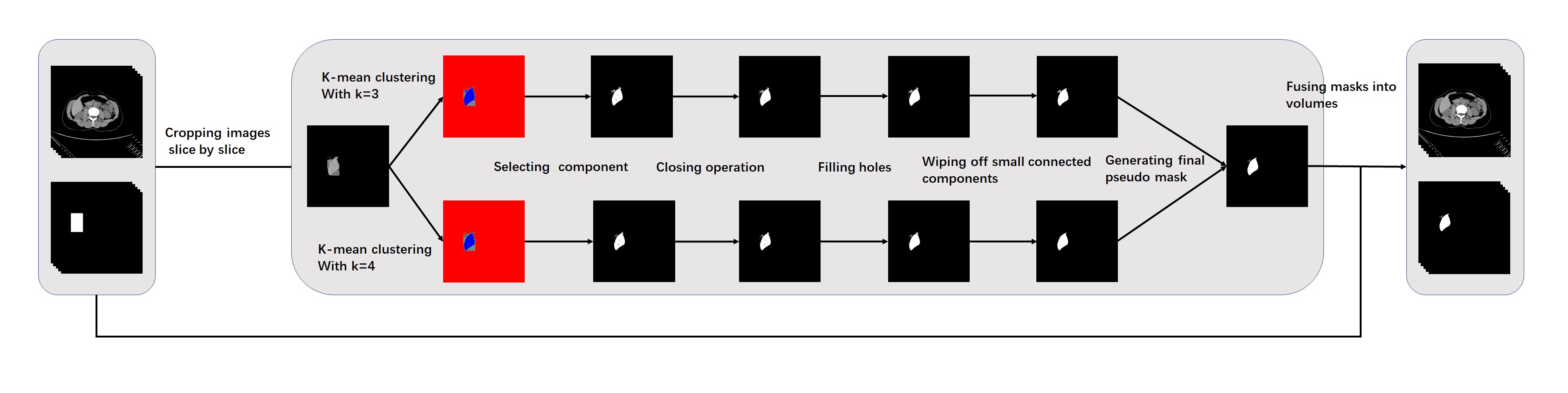}}
	\caption{Procedure of pseudo masks generation.}
	\label{mask-generation}
\end{figure*}

\begin{figure*}[!htp]
	\centering
	\subfigure[]{
		\begin{minipage}[t]{0.15\linewidth}
			\centering
			\includegraphics[width=0.95in]{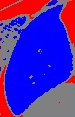}
			%\caption{fig1}
			\label{kmean3}
		\end{minipage}%
	}%
	\subfigure[]{
		\begin{minipage}[t]{0.15\linewidth}
			\centering
			\includegraphics[width=0.95in]{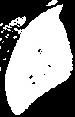}
			%\caption{fig2}
			\label{step2-1}
		\end{minipage}%
	}%
	%这个回车键很重要 \quad也可以
	\subfigure[]{
		\begin{minipage}[t]{0.15\linewidth}
			\centering
			\includegraphics[width=0.95in]{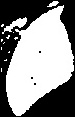}
			%\caption{fig2}
			\label{closing-3}
		\end{minipage}
	}%
	\subfigure[]{
		\begin{minipage}[t]{0.15\linewidth}
			\centering
			\includegraphics[width=0.95in]{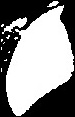}
			%\caption{fig2}
			\label{fillhole-3}
		\end{minipage}
	}%
	\subfigure[]{
		\begin{minipage}[t]{0.15\linewidth}
			\centering
			\includegraphics[width=0.95in]{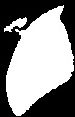}
			%\caption{fig2}
			\label{dc3}
		\end{minipage}
	}%
	\subfigure[]{
		\begin{minipage}[t]{0.15\linewidth}
			\centering
			\includegraphics[width=0.95in]{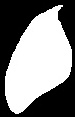}
			%\caption{fig2}
			\label{gt}
		\end{minipage}
	}%
	
	\subfigure[]{
		\begin{minipage}[t]{0.15\linewidth}
			\centering
			\includegraphics[width=0.95in]{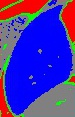}
			%\caption{fig1}
			\label{kmean4}
		\end{minipage}%
	}%
	\subfigure[]{
		\begin{minipage}[t]{0.15\linewidth}
			\centering
			\includegraphics[width=0.95in]{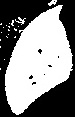}
			%\caption{fig2}
			\label{step2-2}
		\end{minipage}%
	}%
	%这个回车键很重要 \quad也可以
	\subfigure[]{
		\begin{minipage}[t]{0.15\linewidth}
			\centering
			\includegraphics[width=0.95in]{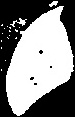}
			%\caption{fig2}
			\label{closing-4}
		\end{minipage}
	}%
	\subfigure[]{
		\begin{minipage}[t]{0.15\linewidth}
			\centering
			\includegraphics[width=0.95in]{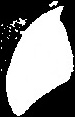}
			%\caption{fig2}
			\label{fillhole-4}
		\end{minipage}
	}%
	\subfigure[]{
		\begin{minipage}[t]{0.15\linewidth}
			\centering
			\includegraphics[width=0.95in]{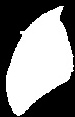}
			%\caption{fig2}
			\label{gc-4}
		\end{minipage}
	}%
	\subfigure[]{
		\begin{minipage}[t]{0.15\linewidth}
			\centering
			\includegraphics[width=0.95in]{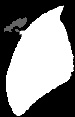}
			%\caption{fig2}
			\label{result}
		\end{minipage}
	}%
	\centering
	\caption{Example generating procedure of pseudo masks generation (only the contents of the bounding box are displayed): (a) result of k-means with $k = 3$, (b) selected component, (c) closing operation, (d) fill all the "holes" (e) removing small connected components, (f) ground truth,(g) result of k-means with $k = 4$, (h) selected component, (i) closing operation, (j) fill all the "holes", (k) removing small connected components, (l) fusing two results into final pseudo mask.}
\end{figure*}

The whole procedure of pseudo mask generation is illustrated in Fig. \ref{mask-generation}. For each slice, we first set the grayscale of all pixels outside the bounding box to 0. Then we perform k-means clustering on the slice (Fig. \ref{kmean3}, Fig. \ref{kmean4}) and choose the class with the second largest area as the foreground and others as the background. Such selection is intuitive and is based on the generation of bounding box labels. In the majority of cases, the region outside the bounding box occupies most of the area of the whole image. After k-means clustering, this outer region, together with a portion of the pixels inside the bounding box, form the cluster with the largest area. Also, bounding box annotations can be regarded as an outward extension of precise labels. Therefore, when such an extension is not tremendous, it is reasonable to choose the cluster with the second largest area as the foreground. Here, the foreground pixels will be labeled as 255 and the background as 0 (Fig. \ref{step2-1}, Fig. \ref{step2-2}). 

Next, we perform a morphological closing operation on the foreground image, i.e., expansion followed by dilation. Such operation has three effects on the image. First, the tiny black holes inside the foreground will be filled in. Second, small sunken areas on the edges of the foreground are filled in so that the entire edge becomes smooth. Finally, some small, densely packed but disconnected foreground areas in the image are fused into larger connected areas (Fig. \ref{closing-3}, Fig. \ref{closing-4}). Inevitably, some black holes still remain inside the foreground image, which may represent noises and lesions inside the organ. Also, there are some remaining small but disconnected white areas. To solve these problems, we directly find out all connected components in the background image and mark the small ones as the foreground to fill the black holes (Fig. \ref{fillhole-3}, Fig. \ref{fillhole-4}). To avoid misclassifying some small connected components that should be in the background as foreground, we set a threshold value of 10 and change the label of a connected component only if its area is smaller than this threshold value. Similarly, we retrieve all the connected components in the foreground image and wipe out the smaller ones following the same method. To reduce false negatives, we set another threshold with a value of 0.01 and mark a connected component as background only if its area is less than one percent of the area of the largest connected component (Fig. \ref{dc3}, Fig. \ref{gc-4}).

The k-means algorithm can roughly classify an image into several clusters, but through experiments, we found that it is very sensitive to the input image and the value of $k$. When the input image is too small and different organs and tissues are not clearly demarcated, or the value of $k$ is too small, the generated foreground will contain more false positives. And when $k$ is too large, the generated foreground will be smaller than the reality. To address this problem, we fuse the masks generated by the k-means algorithm with different values of $k$: when the clustering results of a pixel with different values of $k$ are all belong to the foreground, the label of this pixel is set to 1; when the results all belong to the background, the label is set to 0. If they belong differently with different values of k, the label is set to 0.5 (Fig. \ref{result}). In this study, we choose $k$ values of 3 and 4 for livers as well as 2 and 3 for spleens and kidneys (For kidney images, the left and right kidney are processed separately in each procedure). 

After all the operations above, all CT slices are fused into volumes case by case and prepared to be fed into a CNN model.

\subsection{Deep segmentation model}

\begin{figure}[!htp]
	\centerline{\includegraphics[width=7 in]{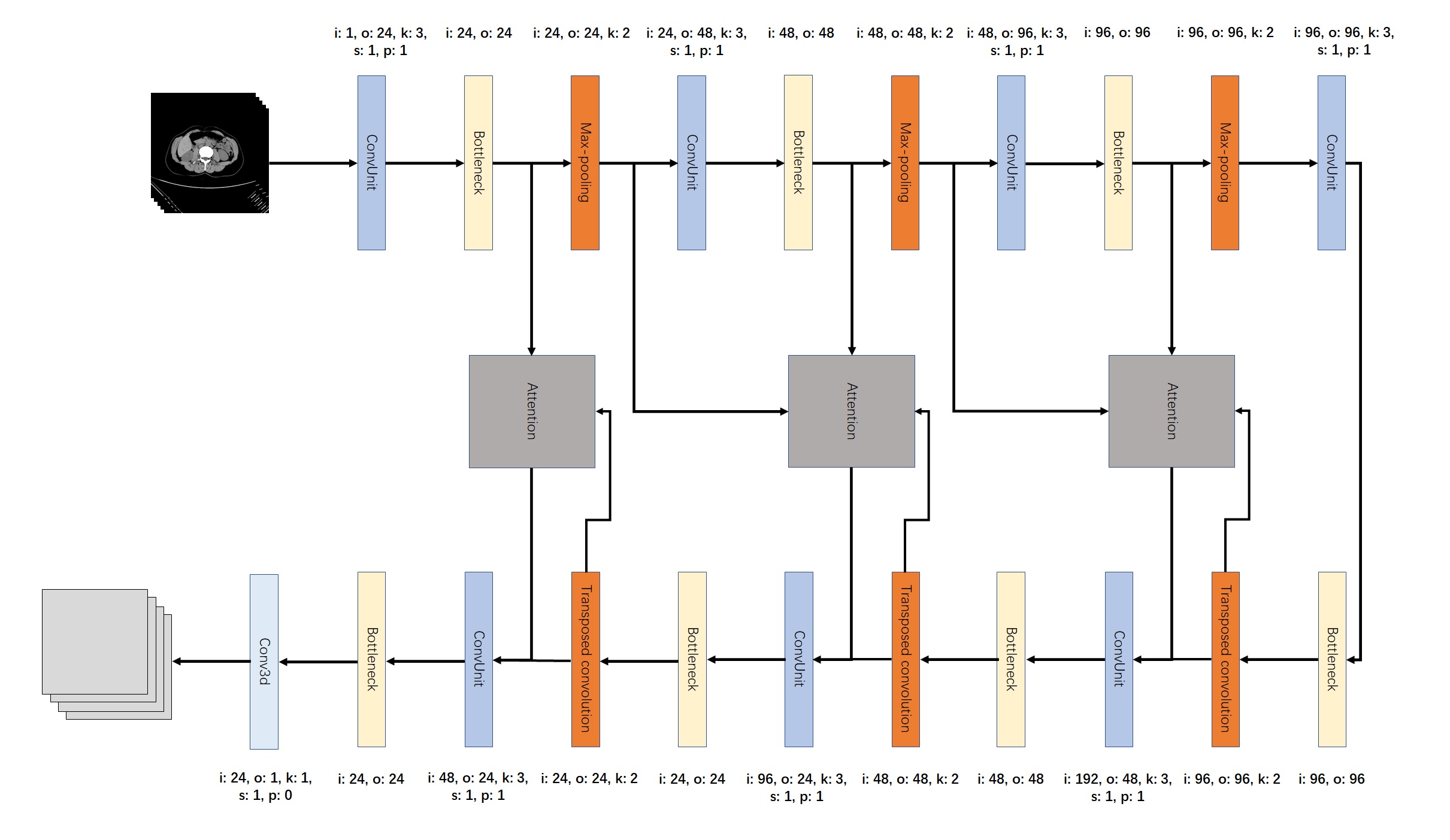}}
	\caption{Structure of the 3D convolutional neural network.}
	\label{unet}
\end{figure}

We propose a new deep segmentation network called BA-Unet (Fig. \ref{unet}). BA-Unet consists of three decoding units and three encoding units, each of which process feature maps at the level of three dimensions. An encoding unit contains a convolution unit to extract features and expand the number of feature maps, a bottleneck block\cite{8578572} to reduce computations while exploiting the
\begin{figure}[h]
	\centerline{\includegraphics[width=3 in]{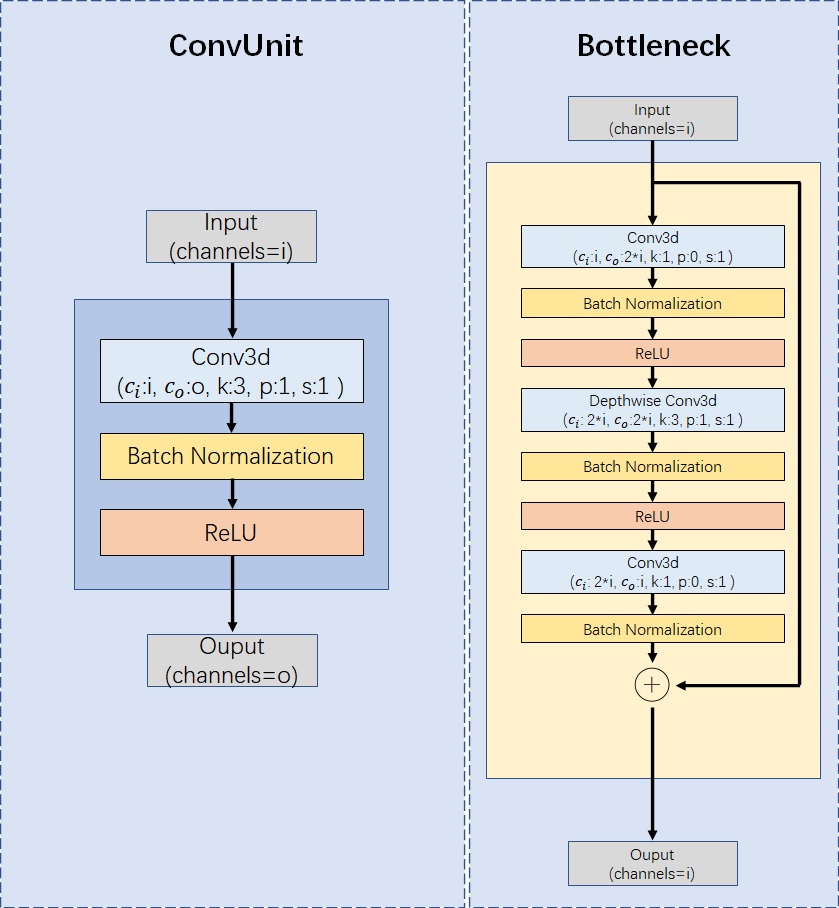}}
	\caption{Structure of a convolution unit and a bottleneck block.}
	\label{c_and_b}
\end{figure}
residual structure\cite{7780459}, and a $2\times 2\times 2$ max-pooling to reduce the dimension of the feature maps. A decoding unit contains a $2 \times 2 \times 2$ transposed convolution to upsample the feature maps, a convolution unit and a bottleneck block to complete the decoding in the current dimension. A convolution unit has a $3 \times 3\times 3$ 3D convolution (Conv3d) with a stride of 1 and padding of 1, which changes the number of features while keeping their dimensions constant. After the convolution, there is a batch normalization\cite{10.5555/3045118.3045167} (BN) and a Rectified Linear Unit (ReLU). A bottleneck block starts with a $1\times 1 \times 1$ Conv3d to reduce the number of feature maps. After a $3 \times 3\times 3$ Conv3d with a step size of 1 and a padding of 1, another $1\times 1 \times 1$ convolution restores the number of feature maps to the initial one. Finally, the output is connected with the input of the bottleneck block in a manner of short connection. Similarly, each Conv3d is directly followed by a BN and a ReLU except for the last one, where the last ReLU is performed after the short connection (Fig. \ref{c_and_b}).

\begin{figure}[!htp]
	\centerline{\includegraphics[width=3 in]{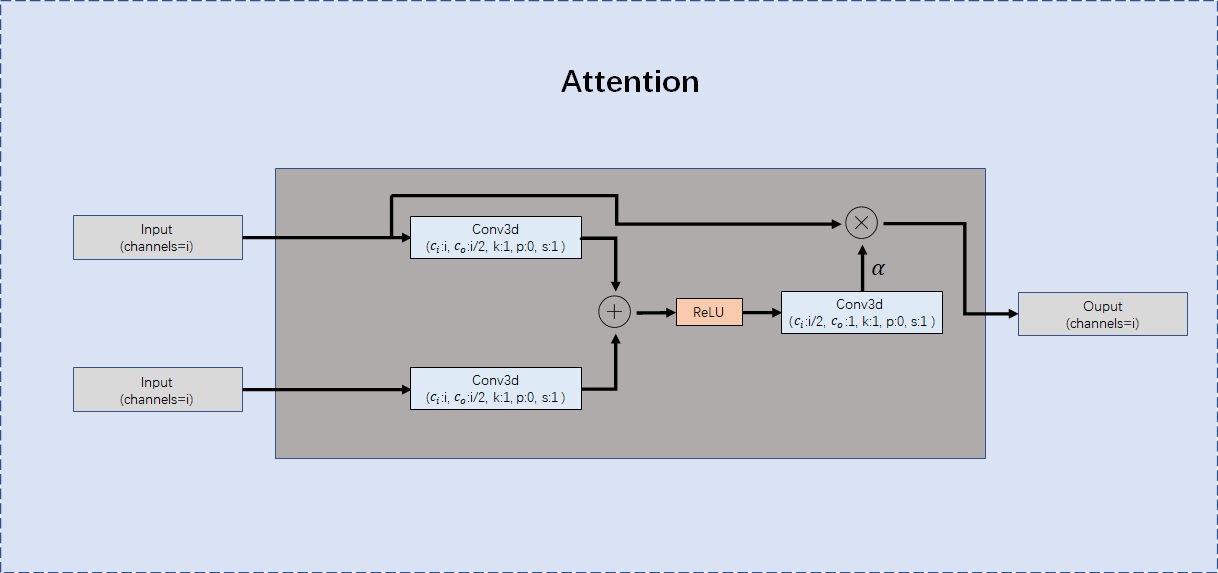}}
	\caption{Structure of an attention block.}
	\label{att}
\end{figure}

Inspired by \cite{oktay2018attention}, we also use the attention blocks to replace the skip connections between the encoder and decoder (Fig. \ref{att}). However, differently from the vanilla attention-Unet\cite{oktay2018attention}, the BA-Unet also incorporates the input of the encoding unit in the current dimension, i.e., the result of the max-pooling in the previous encoding unit, as lower-level features. To predict the target boundaries more accurately, we let the attention blocks additionally consult features from the higher-level encoder to optimize the features in the decoder. After another convolution unit and a bottleneck block, the network ends with a $1\times 1\times 1$ Conv3d to complete the final decoding and generates a probability map by a sigmoid function, which represents the predictions for all voxels.

\section{Experiments}
\label{sec-4} 
We implemented all the experiments with Pytorch. They were run on a Ubuntu 18.04 machine with 32 GB memory and a single NVIDIA RTX 3090 GPU with 24GB of video memory.

\subsection{Data preprocessing}

At the beginning of the experiment, we performed a series of processes on the data. First, the pixel intensity ranges of all images were adjusted to the corresponding Hounsfield Unit intervals, which were (-60, 140) for livers, (-115,185) for spleens, and (-95,155) for kidneys. After that, all images were normalized to [0,1] by setting intensity which was greater than the upper bound to 1 and less than the lower bound to 0. Due to the limitation of video memory, we extracted only the slices occupied by target organs and resized the volumes to a shape of $40 \times 512 \times 512$ for liver and kidneys and $31 \times 512 \times 512$ for spleen.
\begin{figure}[!htp]
	\centering
	
	\subfigure[]{
		\begin{minipage}[t]{0.35\linewidth}
			\centering
			\includegraphics[width=2in]{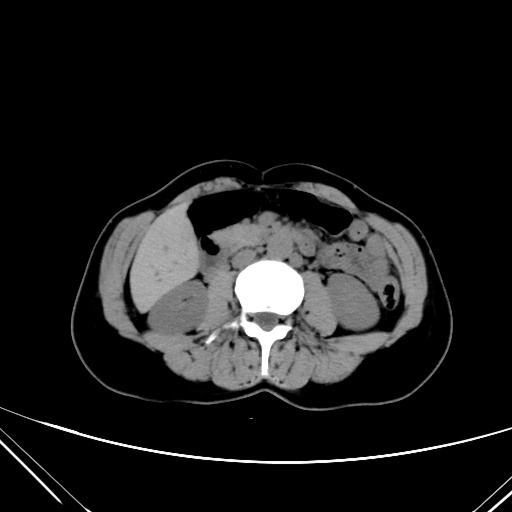}
			%\caption{fig1}
		\end{minipage}%
	}%
	\subfigure[]{
		\begin{minipage}[t]{0.35\linewidth}
			\centering
			\includegraphics[width=2in]{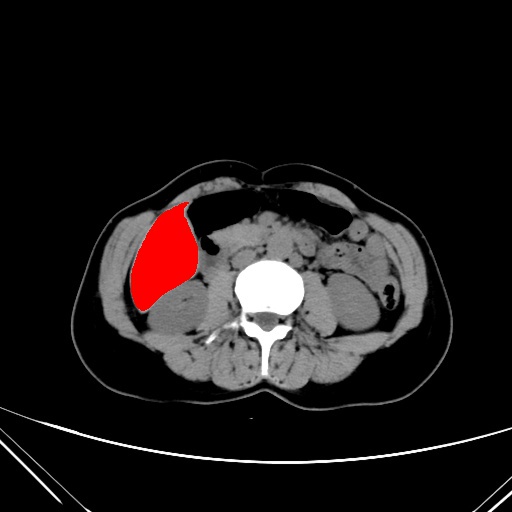}
			%\caption{fig2}
		\end{minipage}%
	}%
	%这个回车键很重要 \quad也可以
	
	\subfigure[]{
		\begin{minipage}[t]{0.35\linewidth}
			\centering
			\includegraphics[width=2in]{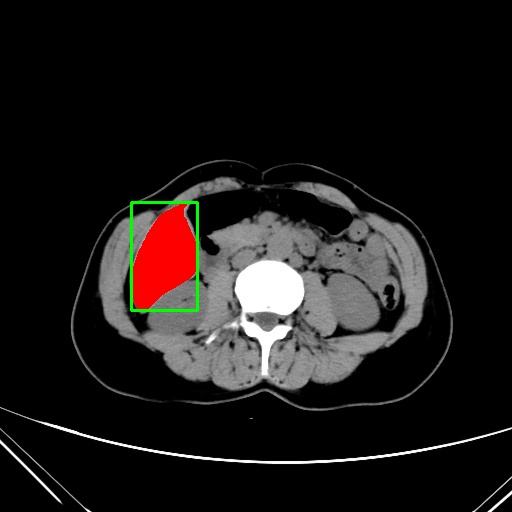}
			%\caption{fig2}
		\end{minipage}
	}%
	\subfigure[]{
		\begin{minipage}[t]{0.35\linewidth}
			\centering
			\includegraphics[width=2in]{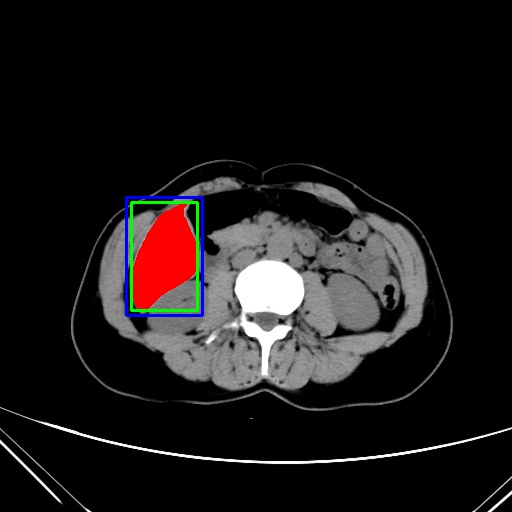}
			%\caption{fig2}
		\end{minipage}
	}%
	
	\centering
	\caption{Example generation of bounding box annotations: (a) original CT slice, (b) pixel-wise ground truth, (c) bounding rectangle (green) and (d) bounding box label (blue). }
	\label{bbox}
\end{figure}
Since the datasets all contain only pixel-wise annotations rather than bounding boxes, we also need to generate the bounding box labels (Fig. \ref{bbox}). For images and annotations in training sets, we extended the bounding rectangle of all ground truth labels of each CT slice by 5 pixels to get the bounding box annotation so that the Dice scores of the bounding box annotations are 66.60\%, 69.12\% and 63.02\% for liver, kidneys, and spleen, respectively. Please note that the bounding box labels are only used in the training phase to generate the initial pseudo masks. 

\subsection{Parameters setting and training details}
\begin{figure}[!htp]
	\centerline{\includegraphics[width=6.8 in]{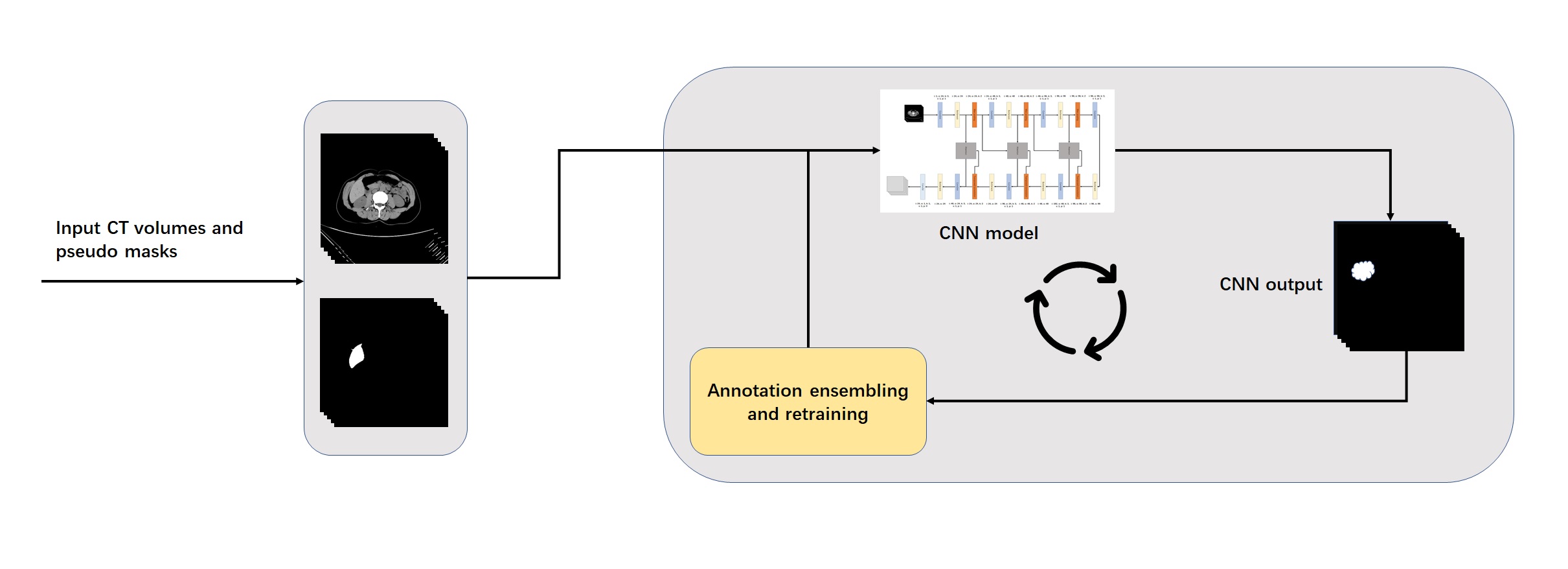}}
	\caption{The training process of the segmentation model. The output of the model is fused with the current labels in a exponential moving average manner and the result is used as the label for the next epoch of training.}
	\label{training}
\end{figure}
The whole training procedure is depicted in Fig. \ref{training}. Following the ideas in \cite{laine2016temporal}, we adopted mechanisms of label ensembling and iterative training. Let $X\in R^{S,W,H}$ denote a training CT volume with its pseudo mask $Y^{0}\in \{0,1\}^{S,W,H}$, $f$ denote the 3D U-net network and $g$ denote the sigmoid function. Then, the probability map output by the network at epoch $t$ is
\begin{equation}	
	y^{t} = g(f^{t}(X))= \frac{1}{1+e^{-f^{t}(X)}}
\end{equation} 
Let $Y^{t}$ denote the cumulative ensemble labels at the beginning of epoch $t$. Then, $Y^{t+1}$ can be calculated by exponential moving average as (\ref{Yt})
\begin{equation}	
	\label{Yt}
	Y^{t+1} = (1-\alpha) Y^{t} + \alpha y^{t}
\end{equation} 
where $\alpha$ is a hyperparameter to control the extent to which the current output $y^{t}$ is incorporated into the historical ensembles. In the experiments, we set $\alpha$ to be 0.1.

We deployed a dice loss in the experiments, which can be written as   
\begin{equation}
	loss = 1-2\times \frac{\sum_{i\in I}Y_{i}y_{i} + \epsilon}{\sum_{i\in I}Y_{i}+\sum_{i\in I}y_{i} + \epsilon}
\end{equation} 
where $\epsilon =0.0000001$ is a smoothing constant to make the denominator non-zero.

We performed a 5-fold cross validation for livers and kidneys as well as a 3-fold cross validation for spleens. In the training phase, An Adam\cite{kingma2014adam} optimizer was firstly deployed with an initial learning rate of 0.0001 for the first three epochs. Each batch contained only one training sample due to the limitation of GPU memory \footnote{ We froze all batch normalization layers to make the training work properly with a batch size of one.}.  For the rest 17 epochs, we implemented a stochastic gradient descent optimizer with an exponentially decayed learning rate as (\ref{lr})

\begin{equation}
	\label{lr}
	{learning\ rate} = {learning\ rate}_{initial} \times {decay\ rate}^{\frac{current\ step}{decayed\ step}}
\end{equation} 
where ${learning\ rate}_{initial}$ was 0.001, ${decay\ rate}$ was 0.94 and $decayed\ step$ was 100 in our experiments.

\subsection{Inference and testing}
In the inference phase, only the normalized CT images without annotations were fed to the CNN to generate predictions. To evaluate our method, We binarized the outputs using 0.5 as the threshold and compared them directly with the pixel-wise ground truth annotations.

\subsection{Evaluation metrics}
In order to present the experimental results more convincingly, we adopted several measurement metrics to evaluate the performance of some popular methods and ours. Those metrics included Dice Similarity Coefficient (DSC), Jaccard Similarity Coefficient (Jaccard), Recall, and Precision. Each measurement was translated to a score in the range from 0 (the lowest possible score) to 100 (the perfect result).

\section{Results}
\label{sec-5}	

\subsection{Results of our method}
For kidney segmentation, each validation fold contained 42 CT scans. Table. \ref{result-kits} summaries the results on each fold and the averaged scores for each metric. The final mean result of DSC, Jaccard, Recall and Precision was 91.45$\pm$7.31, 85.01$\pm$10.93, 91.03$\pm$6.86 and 92.22$\pm$8.54, respectively.

Due to the limited total quantities, we only performed a 3-fold cross validation for spleens, and they contained volumes with a number of 13, 14, and 14, respectively. Table. \ref{result-spleen} describes the results on each fold and the averaged scores. The final mean result of DSC, Jaccard, Recall and Precision was 92.11$\pm$5.61, 85.97$\pm$8.17, 91.89$\pm$4.69 and 93.12$\pm$7.28, respectively.

\par $ $

\begin{table}[h]
	\centering
	
	\begin{tabular}{@{}lllll@{}}
		\toprule
		& DSC                                                    & Jaccard                                                 & Recall                                                 & Precision                                               \\ \midrule
		Fold 1 & \begin{tabular}[c]{@{}l@{}}92.72\\ (4.81)\end{tabular} & \begin{tabular}[c]{@{}l@{}}86.78\\ (7.88)\end{tabular}  & \begin{tabular}[c]{@{}l@{}}91.55\\ (6.22)\end{tabular} & \begin{tabular}[c]{@{}l@{}}94.10\\ (4.22)\end{tabular}  \\
		Fold 2 & \begin{tabular}[c]{@{}l@{}}90.03\\ (9.51)\end{tabular} & \begin{tabular}[c]{@{}l@{}}82.99\\ (13.03)\end{tabular} & \begin{tabular}[c]{@{}l@{}}89.53\\ (9.52)\end{tabular} & \begin{tabular}[c]{@{}l@{}}90.83\\ (10.46)\end{tabular} \\
		Fold 3 & \begin{tabular}[c]{@{}l@{}}90.88\\ (7.62)\end{tabular} & \begin{tabular}[c]{@{}l@{}}84.08\\ (11.48)\end{tabular} & \begin{tabular}[c]{@{}l@{}}90.98\\ (5.85)\end{tabular} & \begin{tabular}[c]{@{}l@{}}91.16\\ (9.82)\end{tabular}  \\
		Fold 4 & \begin{tabular}[c]{@{}l@{}}91.92\\ (7.31)\end{tabular} & \begin{tabular}[c]{@{}l@{}}85.79\\ (11.11)\end{tabular} & \begin{tabular}[c]{@{}l@{}}91.59\\ (6.50)\end{tabular} & \begin{tabular}[c]{@{}l@{}}92.70\\ (9.05)\end{tabular}  \\
		Fold 5 & \begin{tabular}[c]{@{}l@{}}91.71\\ (7.29)\end{tabular} & \begin{tabular}[c]{@{}l@{}}85.43\\ (11.15)\end{tabular} & \begin{tabular}[c]{@{}l@{}}91.52\\ (6.20)\end{tabular} & \begin{tabular}[c]{@{}l@{}}92.31\\ (9.16)\end{tabular}  \\ \midrule
		Avg    & \begin{tabular}[c]{@{}l@{}}91.45\\ (7.31)\end{tabular} & \begin{tabular}[c]{@{}l@{}}85.01\\ (10.93)\end{tabular} & \begin{tabular}[c]{@{}l@{}}91.03\\ (6.86)\end{tabular} & \begin{tabular}[c]{@{}l@{}}92.22\\ (8.54)\end{tabular}  \\ \bottomrule
	\end{tabular}
	\caption{Numerical accuracy results for kidneys segmentation. All measurements are reported as fold-averaged (standard variation) [\%]}
	\label{result-kits}
\end{table}
\par $ $

\begin{table}[h]
	\centering
	
	\begin{tabular}{@{}lllll@{}}
		\toprule
		& DSC                                                     & Jaccard                                                 & Recall                                                 & Precision                                               \\ \midrule
		Fold 1 & \begin{tabular}[c]{@{}l@{}}92.95\\ (2.94)\end{tabular}  & \begin{tabular}[c]{@{}l@{}}86.97\\ (5.04)\end{tabular}  & \begin{tabular}[c]{@{}l@{}}90.32\\ (4.88)\end{tabular} & \begin{tabular}[c]{@{}l@{}}95.95\\ (2.94)\end{tabular}  \\
		Fold 2 & \begin{tabular}[c]{@{}l@{}}93.88\\ (3.23)\end{tabular}  & \begin{tabular}[c]{@{}l@{}}88.63\\ (5.57)\end{tabular}  & \begin{tabular}[c]{@{}l@{}}93.25\\ (4.70)\end{tabular} & \begin{tabular}[c]{@{}l@{}}94.75\\ (4.17)\end{tabular}  \\
		Fold 3 & \begin{tabular}[c]{@{}l@{}}89.49\\ (10.66)\end{tabular} & \begin{tabular}[c]{@{}l@{}}82.31\\ (13.90)\end{tabular} & \begin{tabular}[c]{@{}l@{}}92.11\\ (4.49)\end{tabular} & \begin{tabular}[c]{@{}l@{}}88.67\\ (14.73)\end{tabular} \\ \midrule
		Avg    & \begin{tabular}[c]{@{}l@{}}92.11\\ (5.61)\end{tabular}  & \begin{tabular}[c]{@{}l@{}}85.97\\ (8.17)\end{tabular}  & \begin{tabular}[c]{@{}l@{}}91.89\\ (4.69)\end{tabular} & \begin{tabular}[c]{@{}l@{}}93.12\\ (7.28)\end{tabular}  \\ \bottomrule
	\end{tabular}
	\caption{Numerical accuracy results for spleen segmentation. All measurements are reported as fold-averaged (standard variation) [\%]}
	\label{result-spleen}
\end{table}

For liver segmentation, each fold in the 5-fold cross-validation contains 49 images. Table. \ref{result-liver} illustrates the results on each fold and the averaged scores. The final mean result of DSC, Jaccard, Recall and Precision was 95.19$\pm$0.84, 90.83$\pm$1.51, 95.20$\pm$1.06 and 95.20$\pm$1.44, respectively.

In addition, we performed an external validation of liver segmentation. Each trained model was directly validated on 131 CT scans from LiTS dataset. Table. \ref{result-liver} illustrates the results on each fold and the averaged scores. The final mean result of DSC, Jaccard, Recall and Precision was 93.49$\pm$2.17, 88.20$\pm$3.92, 94.11$\pm$2.92 and 93.27$\pm$3.49, respectively.
\par $ $
\begin{table}[!htp]
	\centering
	
	\begin{tabular}{@{}lllll@{}}
		\toprule
		& DSC                                                    & Jaccard                                                & Recall                                                 & Precision                                              \\ \midrule
		Fold 1 & \begin{tabular}[c]{@{}l@{}}95.02\\ (1.05)\end{tabular} & \begin{tabular}[c]{@{}l@{}}90.52\\ (1.88)\end{tabular} & \begin{tabular}[c]{@{}l@{}}94.85\\ (1.93)\end{tabular} & \begin{tabular}[c]{@{}l@{}}95.23\\ (1.68)\end{tabular} \\
		Fold 2 & \begin{tabular}[c]{@{}l@{}}94.97\\ (0.82)\end{tabular} & \begin{tabular}[c]{@{}l@{}}90.44\\ (1.48)\end{tabular} & \begin{tabular}[c]{@{}l@{}}95.12\\ (0.79)\end{tabular} & \begin{tabular}[c]{@{}l@{}}94.84\\ (1.36)\end{tabular} \\
		Fold 3 & \begin{tabular}[c]{@{}l@{}}95.11\\ (0.64)\end{tabular} & \begin{tabular}[c]{@{}l@{}}90.68\\ (1.17)\end{tabular} & \begin{tabular}[c]{@{}l@{}}94.88\\ (0.76)\end{tabular} & \begin{tabular}[c]{@{}l@{}}95.35\\ (1.43)\end{tabular} \\
		Fold 4 & \begin{tabular}[c]{@{}l@{}}95.45\\ (1.09)\end{tabular} & \begin{tabular}[c]{@{}l@{}}91.32\\ (1.97)\end{tabular} & \begin{tabular}[c]{@{}l@{}}95.41\\ (0.92)\end{tabular} & \begin{tabular}[c]{@{}l@{}}95.50\\ (1.80)\end{tabular} \\
		Fold 5 & \begin{tabular}[c]{@{}l@{}}95.40\\ (0.58)\end{tabular} & \begin{tabular}[c]{@{}l@{}}91.21\\ (1.06)\end{tabular} & \begin{tabular}[c]{@{}l@{}}95.73\\ (0.89)\end{tabular} & \begin{tabular}[c]{@{}l@{}}95.09\\ (0.93)\end{tabular} \\ \midrule
		Avg    & \begin{tabular}[c]{@{}l@{}}95.19\\ (0.84)\end{tabular} & \begin{tabular}[c]{@{}l@{}}90.83\\ (1.51)\end{tabular} & \begin{tabular}[c]{@{}l@{}}95.20\\ (1.06)\end{tabular} & \begin{tabular}[c]{@{}l@{}}95.20\\ (1.44)\end{tabular} \\ \bottomrule
	\end{tabular}
	\caption{Numerical accuracy results for liver segmentation. All measurements are reported as fold-averaged (standard variation) [\%]}
	\label{result-liver}
\end{table}
\par $ $
\begin{table}[!htp]
	\centering
	
	\begin{tabular}{@{}lllll@{}}
		\toprule
		& DSC                                                    & Jaccard                                                & Recall                                                 & Precision                                              \\ \midrule
		Fold 1 & \begin{tabular}[c]{@{}l@{}}93.50\\ (2.30)\end{tabular} & \begin{tabular}[c]{@{}l@{}}87.88\\ (5.04)\end{tabular} & \begin{tabular}[c]{@{}l@{}}94.38\\ (2.82)\end{tabular} & \begin{tabular}[c]{@{}l@{}}92.78\\ (3.81)\end{tabular} \\
		Fold 2 & \begin{tabular}[c]{@{}l@{}}93.60\\ (2.02)\end{tabular} & \begin{tabular}[c]{@{}l@{}}88.04\\ (3.45)\end{tabular} & \begin{tabular}[c]{@{}l@{}}94.37\\ (2.82)\end{tabular} & \begin{tabular}[c]{@{}l@{}}92.30\\ (3.23)\end{tabular} \\
		Fold 3 & \begin{tabular}[c]{@{}l@{}}93.78\\ (2.18)\end{tabular} & \begin{tabular}[c]{@{}l@{}}88.36\\ (3.70)\end{tabular} & \begin{tabular}[c]{@{}l@{}}93.98\\ (2.99)\end{tabular} & \begin{tabular}[c]{@{}l@{}}93.72\\ (3.45)\end{tabular} \\
		Fold 4 & \begin{tabular}[c]{@{}l@{}}93.79\\ (2.16)\end{tabular} & \begin{tabular}[c]{@{}l@{}}88.38\\ (3.66)\end{tabular} & \begin{tabular}[c]{@{}l@{}}93.90\\ (2.93)\end{tabular} & \begin{tabular}[c]{@{}l@{}}93.81\\ (3.45)\end{tabular} \\
		Fold 5 & \begin{tabular}[c]{@{}l@{}}93.76\\ (2.21)\end{tabular} & \begin{tabular}[c]{@{}l@{}}88.33\\ (3.74)\end{tabular} & \begin{tabular}[c]{@{}l@{}}93.94\\ (3.03)\end{tabular} & \begin{tabular}[c]{@{}l@{}}93.72\\ (3.51)\end{tabular} \\ \midrule
		Avg    & \begin{tabular}[c]{@{}l@{}}93.49\\ (2.17)\end{tabular} & \begin{tabular}[c]{@{}l@{}}88.20\\ (3.92)\end{tabular} & \begin{tabular}[c]{@{}l@{}}94.11\\ (2.92)\end{tabular} & \begin{tabular}[c]{@{}l@{}}93.27\\ (3.49)\end{tabular} \\ \bottomrule
	\end{tabular}
	\caption{Numerical accuracy results for liver segmentation on external validation set. All measurements are reported as fold-averaged (standard variation) [\%]}
	\label{result-lits}
\end{table}

\subsection{Additional experiments}
Following the same data set division as before, we also experimented with the other three methods in a cross-validation manner. They are training BA-Unet with the original bounding box (BA-Unet$_{bbox}$), segmenting directly with GrabCut, and training BA-Unet with the results of GrabCut (U-net$_{GC}$). For each method, we calculated the mean DSC and Jaccard score with their standard variations. It is necessary to mention that we found GrabCut was so struggled in segmenting the spleen data used in this study that for more than half of the scans, the results were totally blank. Thus we only represent the results of BA-Unet$_{bbox}$. 
\par $ $
\begin{table}[!htp]
	\centering
	\setlength{\tabcolsep}{0.05mm}{
		\begin{tabular}{@{}lllll@{}}
			\toprule
			& BA-Unet$_{Bbox} $                                                      & GrabCut                                                              & BA-Unet$_{GC}$                                                           & Ours                                                                         \\ \midrule
			Fold 1 & \begin{tabular}[c]{@{}l@{}}72.19(5.65)/\\ 56.77(6.69)\end{tabular} & \begin{tabular}[c]{@{}l@{}}78.07(18.00)/\\ 66.95(19.90)\end{tabular} & \begin{tabular}[c]{@{}l@{}}84.54(7.27)/\\ 73.87(10.39)\end{tabular}  & \begin{tabular}[c]{@{}l@{}}92.72(4.81)/\\ 86.78(7.88)\end{tabular}           \\
			Fold 2 & \begin{tabular}[c]{@{}l@{}}70.83(9.10)/\\ 55.53(9.94)\end{tabular} & \begin{tabular}[c]{@{}l@{}}73.97(21.18)/\\ 62.38(22.21)\end{tabular} & \begin{tabular}[c]{@{}l@{}}79.18(16.02)/\\ 67.83(17.67)\end{tabular} & \begin{tabular}[c]{@{}l@{}}90.03(9.51)/\\ 82.99(13.03)\end{tabular}          \\
			Fold 3 & \begin{tabular}[c]{@{}l@{}}71.23(8.23)/\\ 55.90(9.22)\end{tabular} & \begin{tabular}[c]{@{}l@{}}76.76(16.95)/\\ 64.93(19.40)\end{tabular} & \begin{tabular}[c]{@{}l@{}}80.05(13.23)/\\ 68.54(16.50)\end{tabular} & \begin{tabular}[c]{@{}l@{}}90.88(7.62)/\\ 84.08(11.48)\end{tabular}          \\
			Fold 4 & \begin{tabular}[c]{@{}l@{}}71.47(7.98)/\\ 56.15(8.88)\end{tabular} & \begin{tabular}[c]{@{}l@{}}74.59(21.28)/\\ 63.42(23.45)\end{tabular} & \begin{tabular}[c]{@{}l@{}}82.47(12.50)/\\ 71.81(15.54)\end{tabular} & \begin{tabular}[c]{@{}l@{}}91.92(7.31)/\\ 85.79(11.11)\end{tabular}          \\
			Fold 5 & \begin{tabular}[c]{@{}l@{}}71.75(7.63)/\\ 56.44(8.43)\end{tabular} & \begin{tabular}[c]{@{}l@{}}77.76(16.97)/\\ 66.28(19.46)\end{tabular} & \begin{tabular}[c]{@{}l@{}}82.51(11.13)/\\ 71.62(14.85)\end{tabular} & \begin{tabular}[c]{@{}l@{}}91.71(7.29)/\\ 85.43(11.15)\end{tabular}          \\ \midrule
			Avg    & \begin{tabular}[c]{@{}l@{}}71.49(7.72)/\\ 56.16(\textbf{8.63})\end{tabular} & \begin{tabular}[c]{@{}l@{}}76.23(16.88)/\\ 64.79(18.88)\end{tabular} & \begin{tabular}[c]{@{}l@{}}81.75(12.03)/\\ 70.73(14.99)\end{tabular} & \begin{tabular}[c]{@{}l@{}}\textbf{91.45(7.31)}/\\ \textbf{85.01}(10.93)\end{tabular} \\ \bottomrule
	\end{tabular}}
	\caption{Comparison of different methods on kidneys segmentation. All measurements are reported as fold-averaged (standard variation) DSC and Jaccard scores [\%]. The best scores are marked in bold.
	}
	\label{compare-kidneys}
\end{table}

\begin{figure}[h]
	\centering
	\subfigure[]{
		\begin{minipage}[t]{0.18\linewidth}
			\centering
			\includegraphics[width=1.3in]{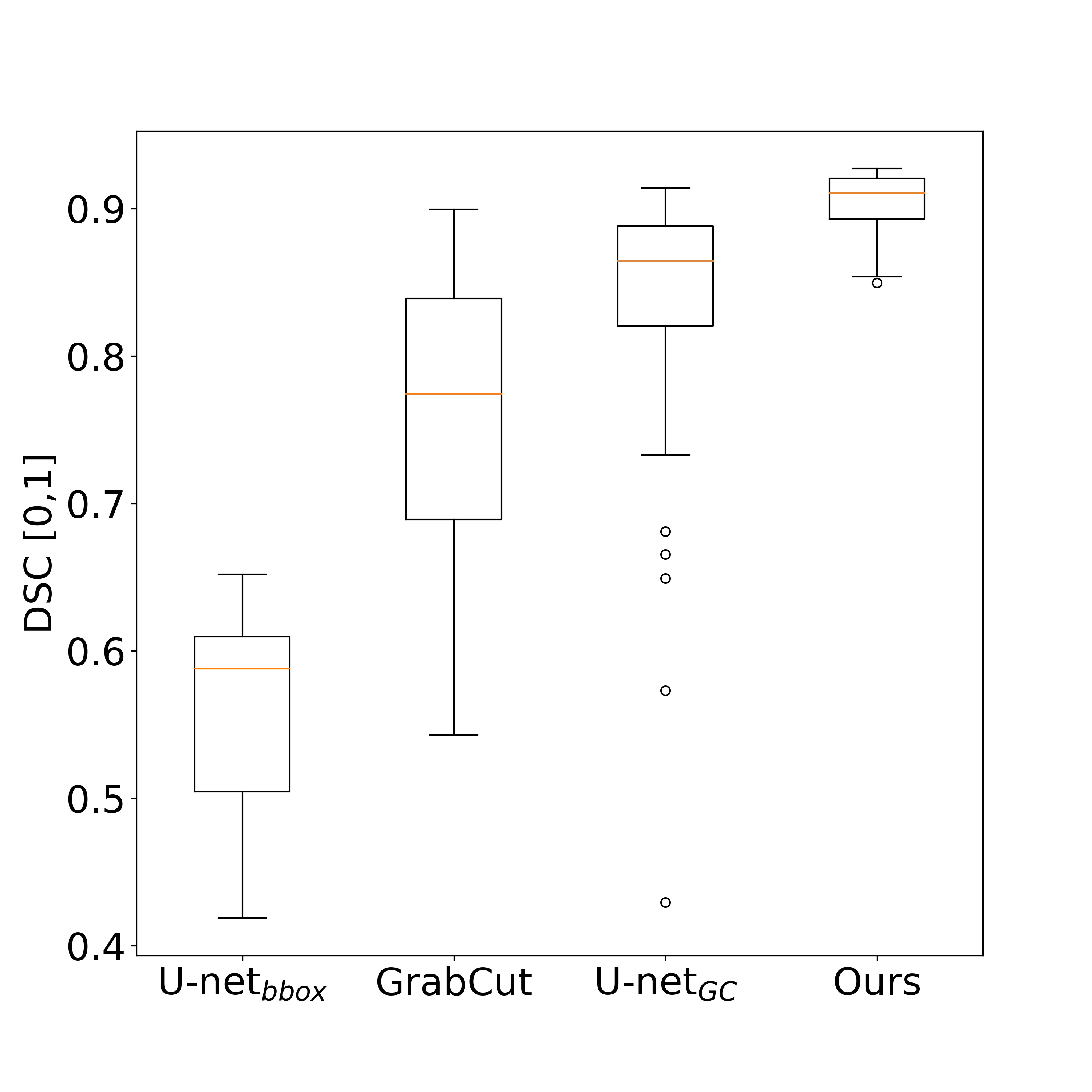}
			%\caption{fig1}
			\label{liver_box_1}
		\end{minipage}%
	}%
	\subfigure[]{
		\begin{minipage}[t]{0.18\linewidth}
			\centering
			\includegraphics[width=1.3in]{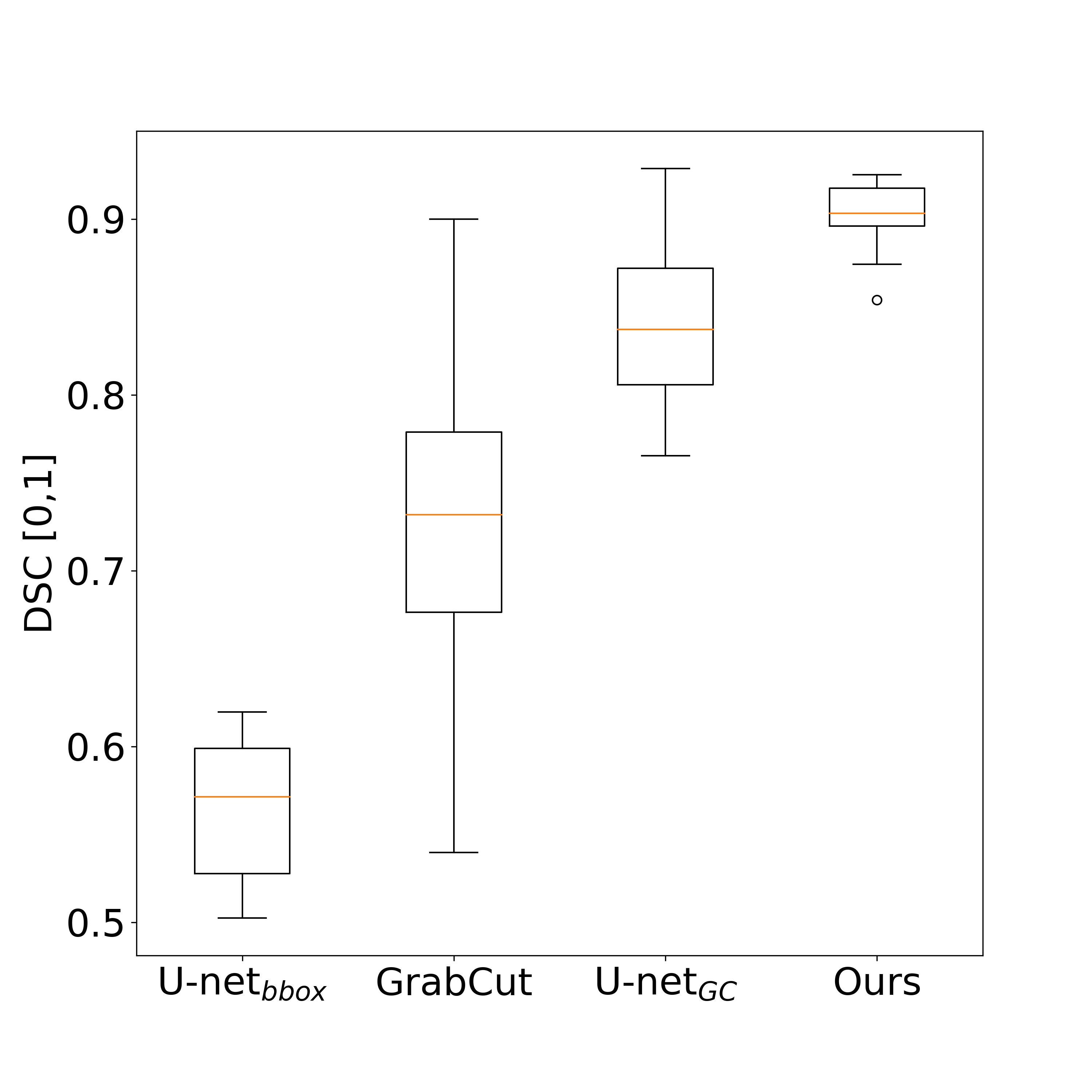}
			%\caption{fig2}
			\label{liver_box_2}
		\end{minipage}
	}%
	\subfigure[]{
		\begin{minipage}[t]{0.18\linewidth}
			\centering
			\includegraphics[width=1.3in]{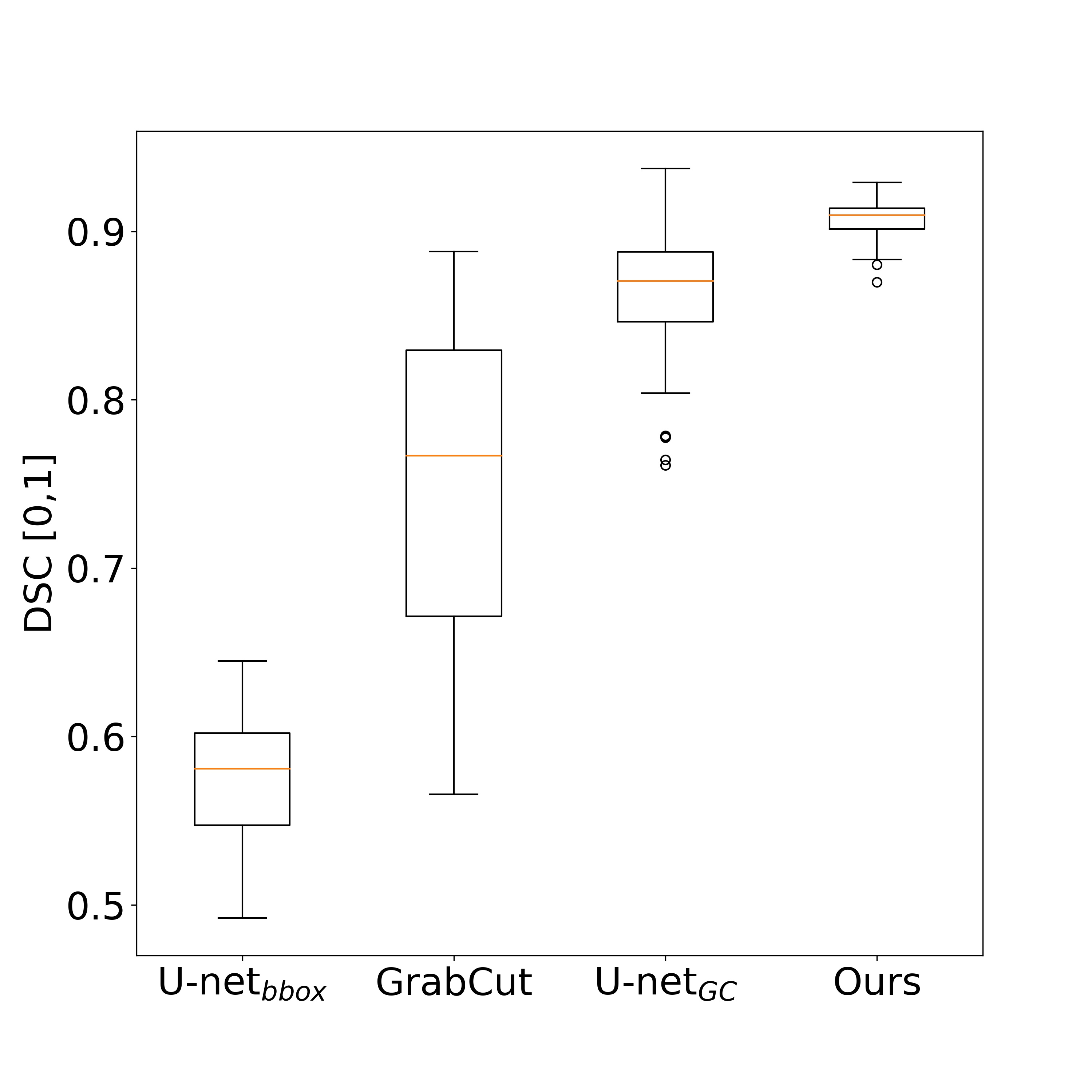}
			%\caption{fig2}
			\label{liver_box_3}
		\end{minipage}
	}%
	\subfigure[]{
		\begin{minipage}[t]{0.18\linewidth}
			\centering
			\includegraphics[width=1.3in]{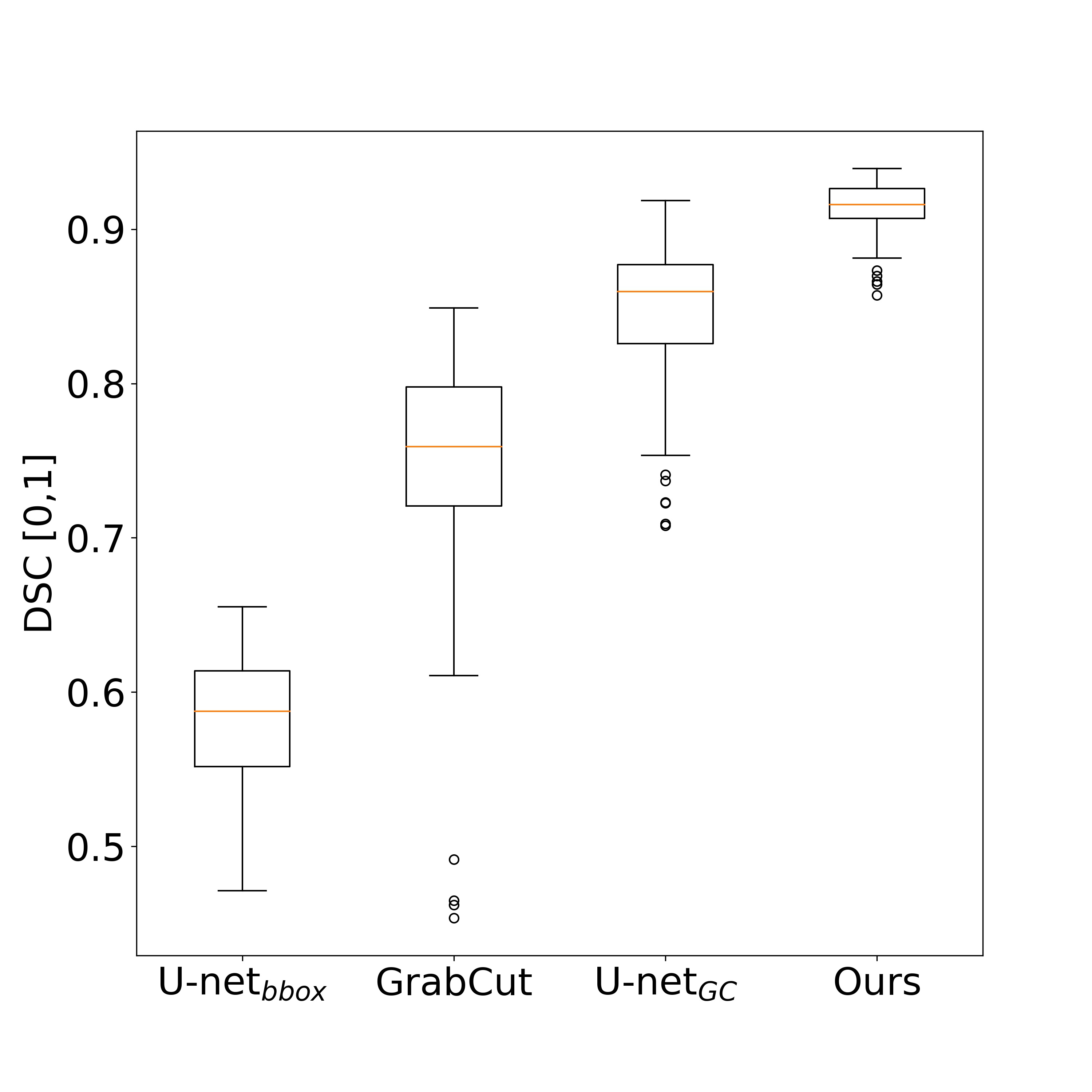}
			%\caption{fig2}
			\label{liver_box_4}
		\end{minipage}
	}%
	\subfigure[]{
		\begin{minipage}[t]{0.18\linewidth}
			\centering
			\includegraphics[width=1.3in]{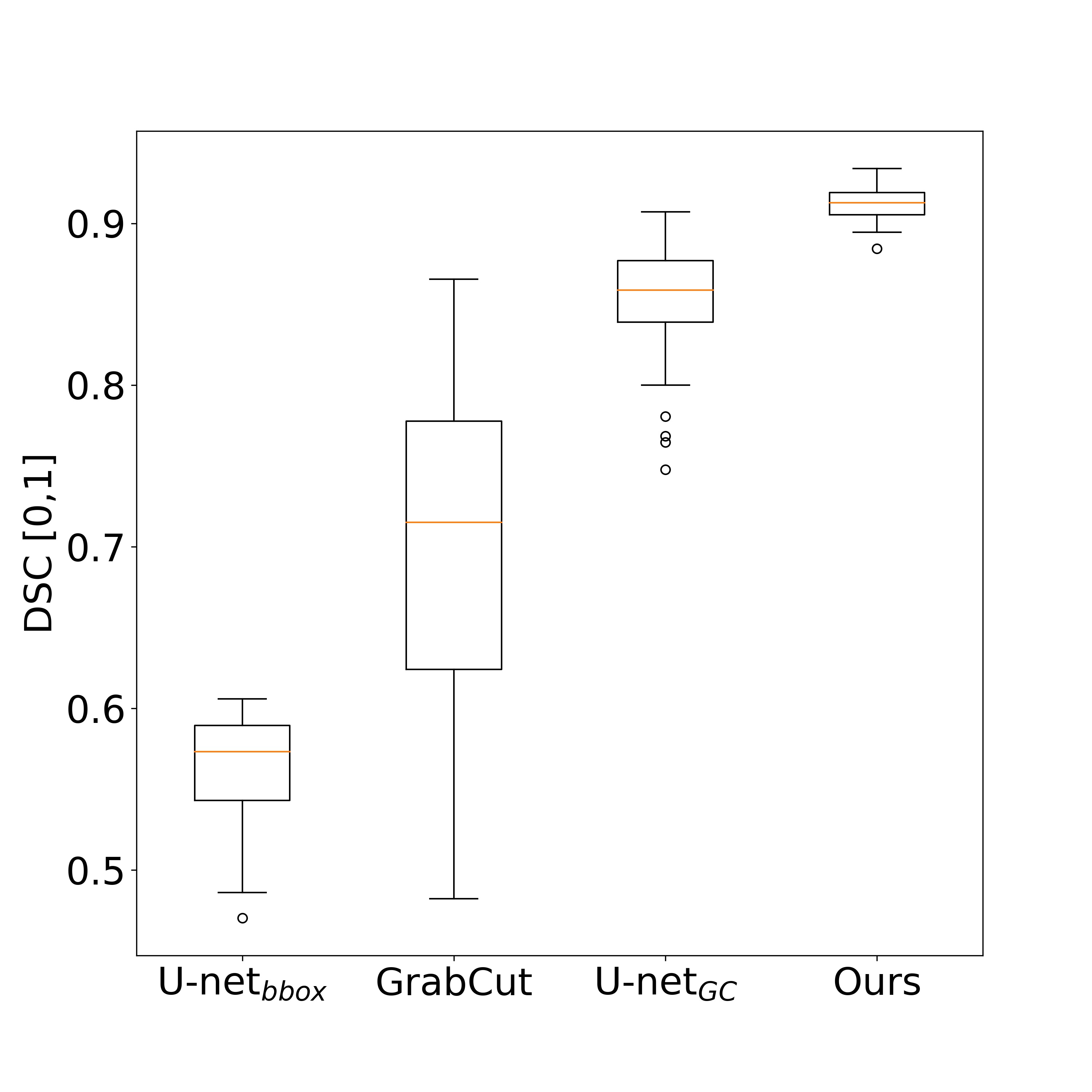}
			%\caption{fig2}
			\label{liver_box_5}
		\end{minipage}%
	}%
	%这个回车键很重要 \quad也可以
	
	\subfigure[]{
		\begin{minipage}[t]{0.18\linewidth}
			\centering
			\includegraphics[width=1.3 in]{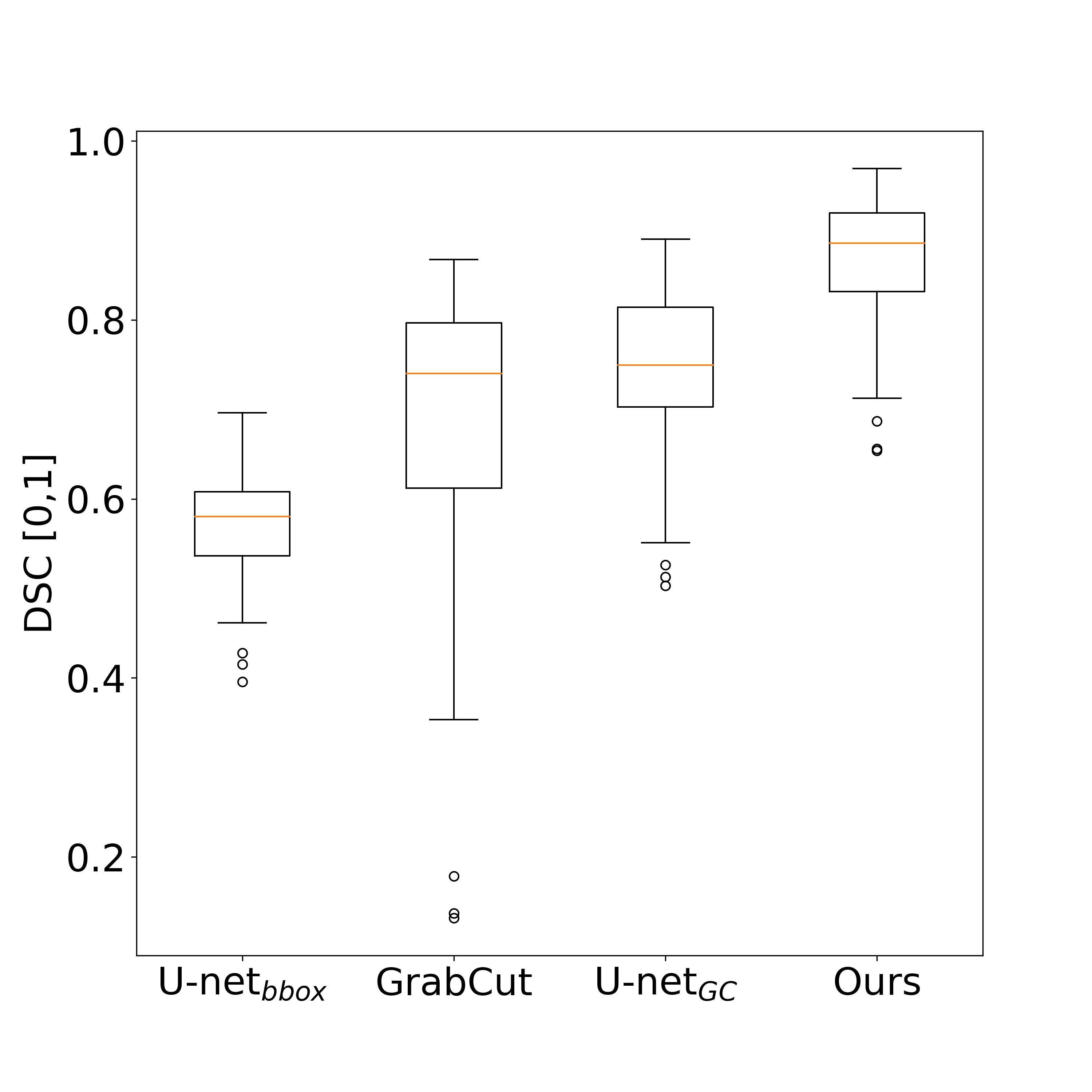}
			%\caption{fig2}
			\label{kits_box_1}
		\end{minipage}
	}%
	\subfigure[]{
		\begin{minipage}[t]{0.18\linewidth}
			\centering
			\includegraphics[width=1.3 in]{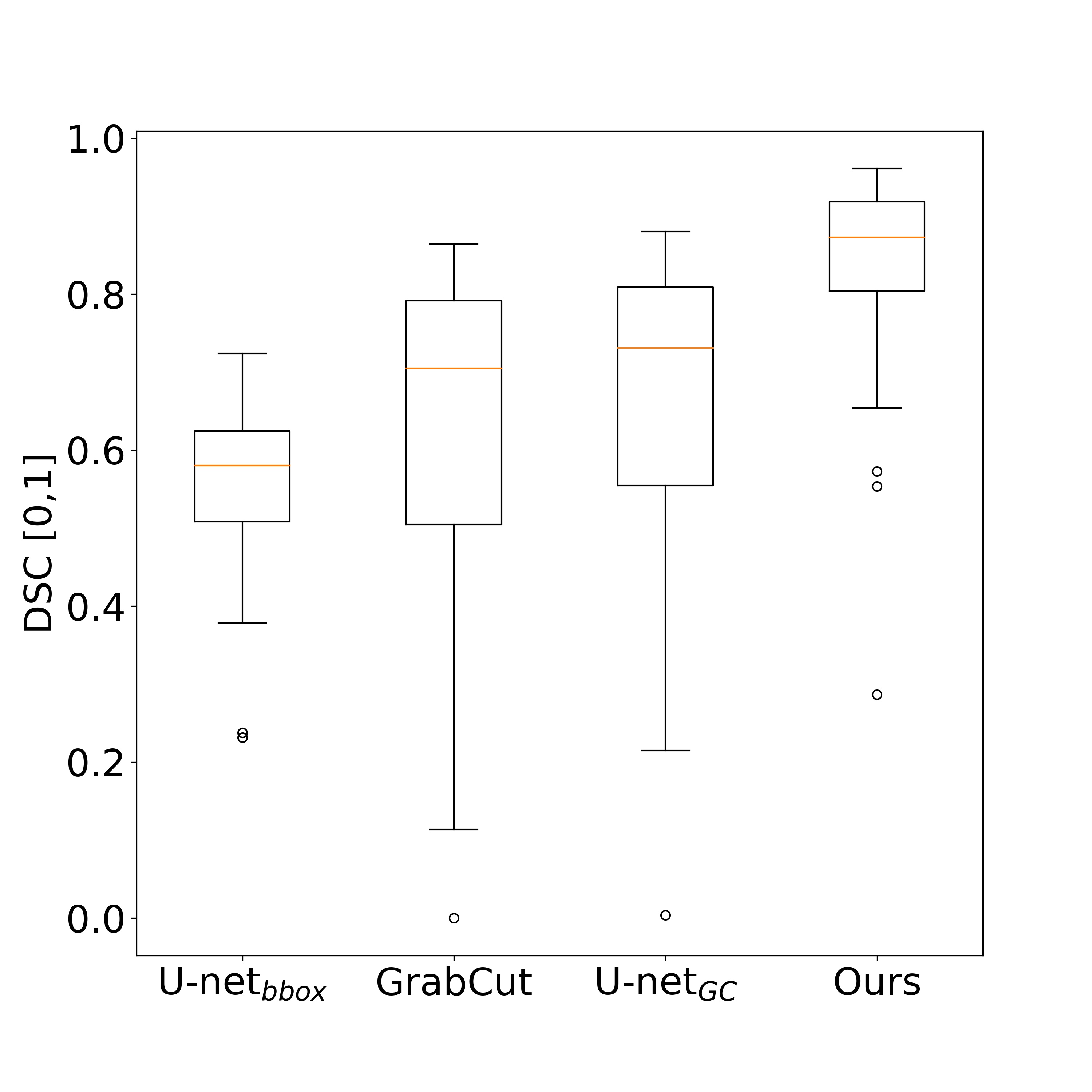}
			%\caption{fig2}
			\label{kits_box_2}
		\end{minipage}
	}%
	\subfigure[]{
		\begin{minipage}[t]{0.18\linewidth}
			\centering
			\includegraphics[width=1.3 in]{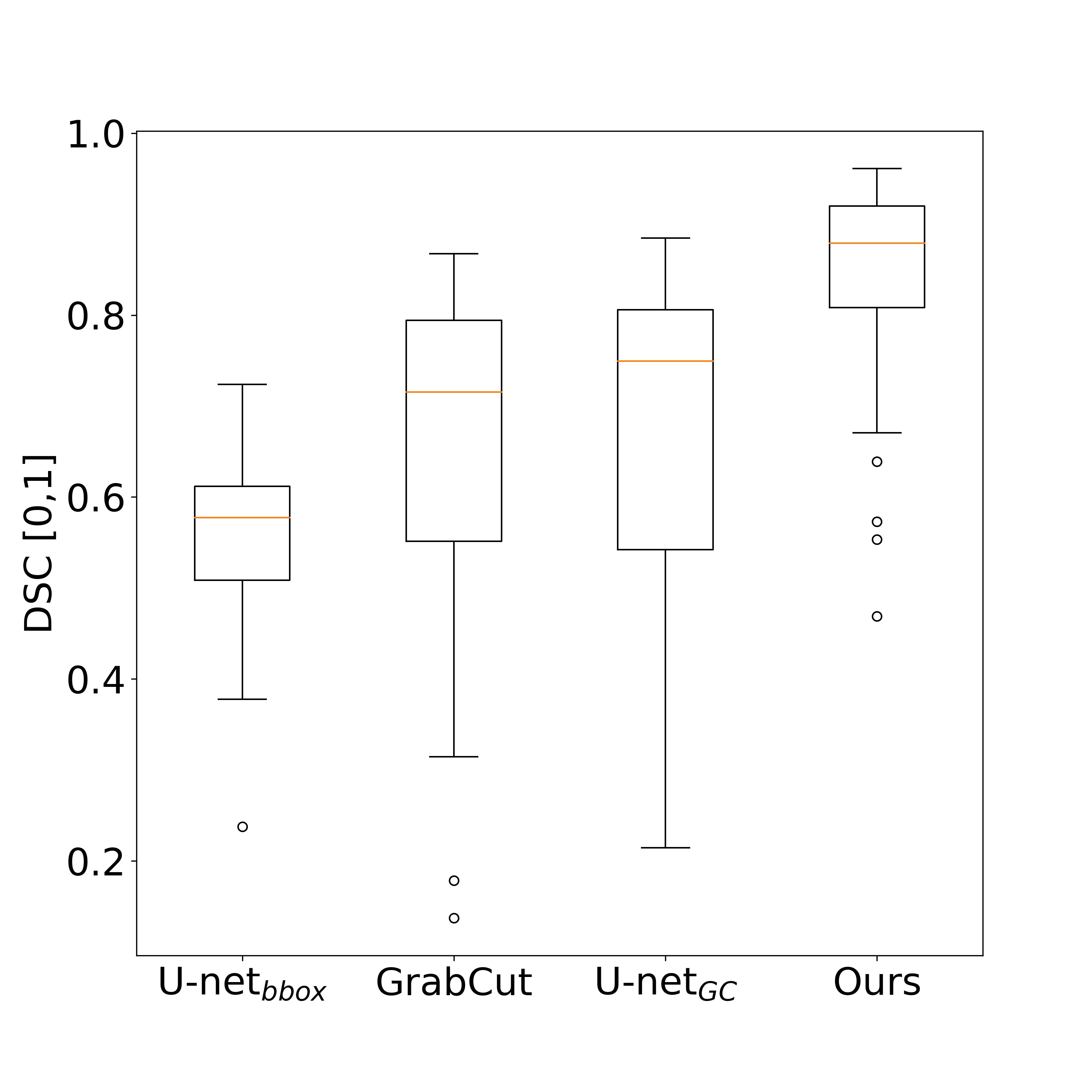}
			%\caption{fig2}
			\label{kits_box_3}
		\end{minipage}
	}%
	\subfigure[]{
		\begin{minipage}[t]{0.18\linewidth}
			\centering
			\includegraphics[width=1.3 in]{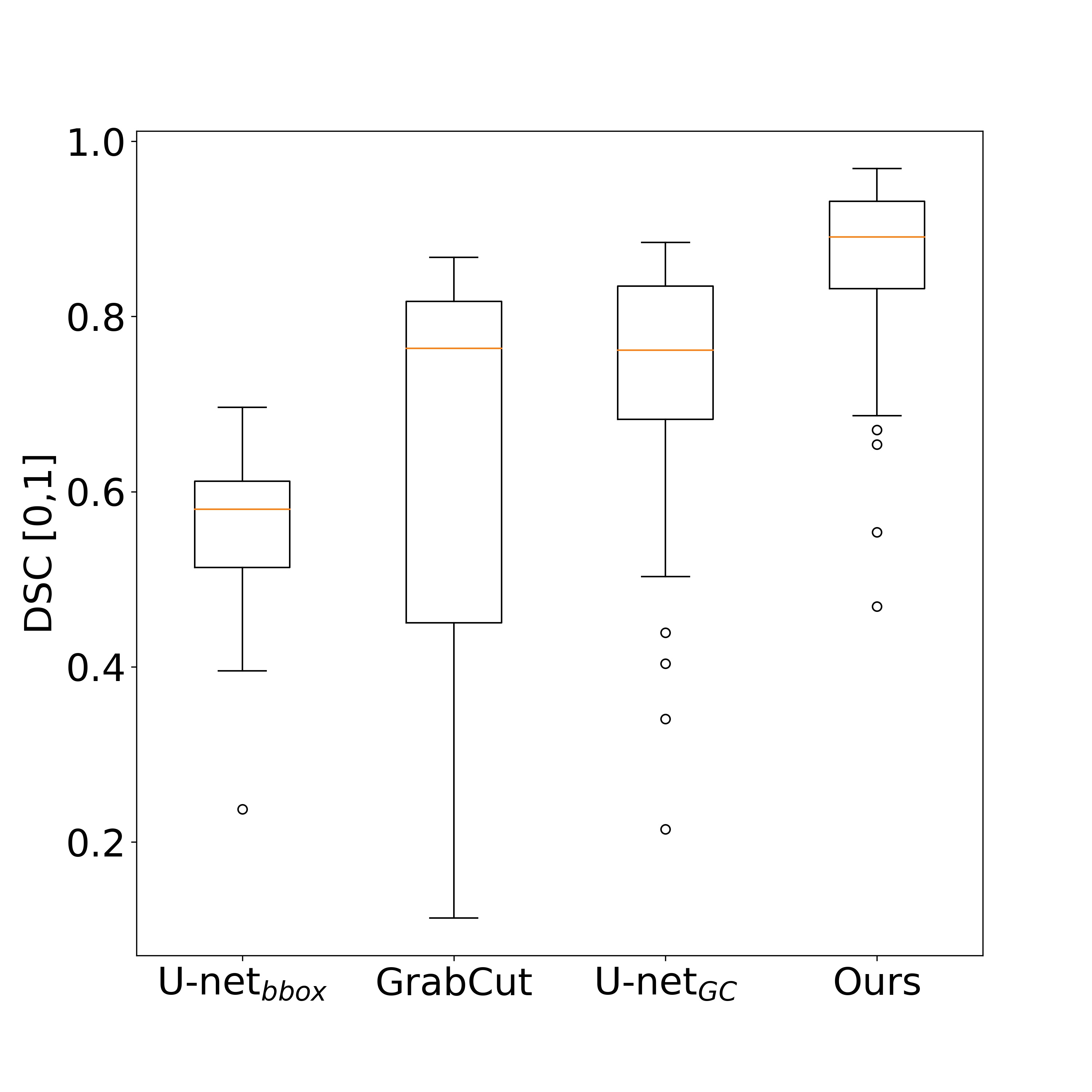}
			%\caption{fig2}
			\label{kits_box_4}
		\end{minipage}
	}%
	\subfigure[]{
		\begin{minipage}[t]{0.18 \linewidth}
			\centering
			\includegraphics[width=1.3 in]{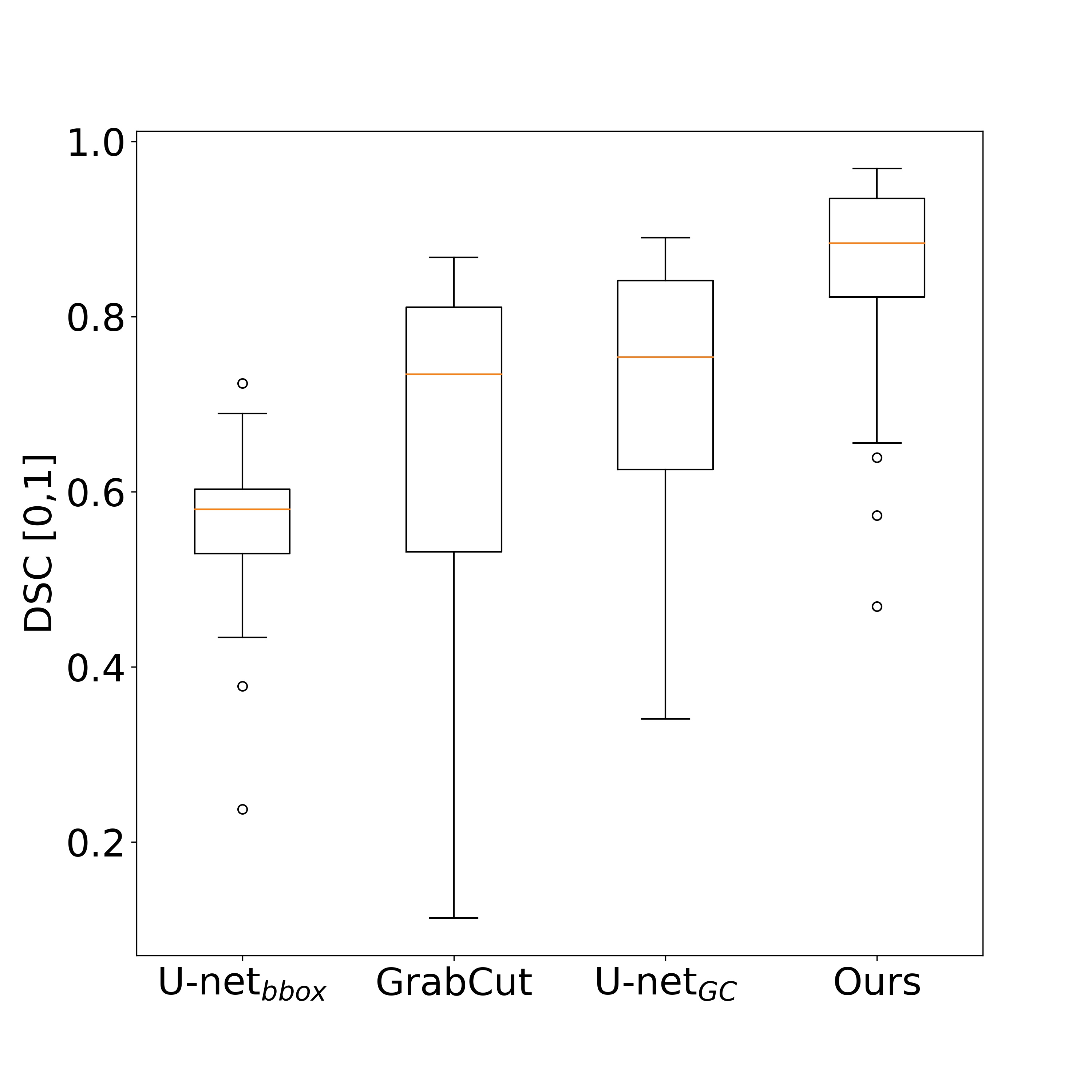}
			%\caption{fig2}
			\label{kits_box_5}
		\end{minipage}
	}%
	
	\subfigure[]{
		\begin{minipage}[t]{0.18\linewidth}
			\centering
			\includegraphics[width=1.3 in]{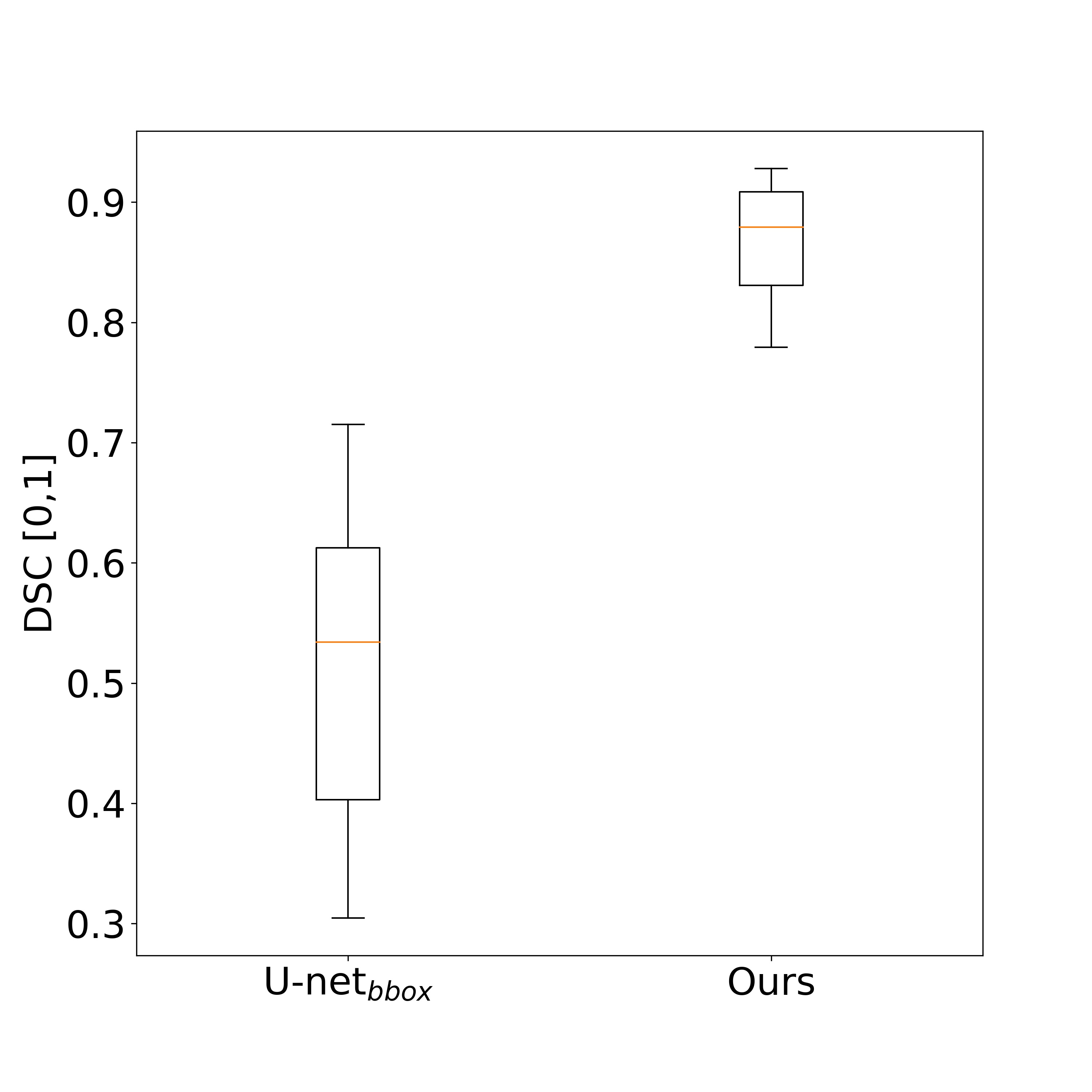}
			%\caption{fig2}
			\label{spleen_box_1}
		\end{minipage}
	}%
	\subfigure[]{
		\begin{minipage}[t]{0.18\linewidth}
			\centering
			\includegraphics[width=1.3 in]{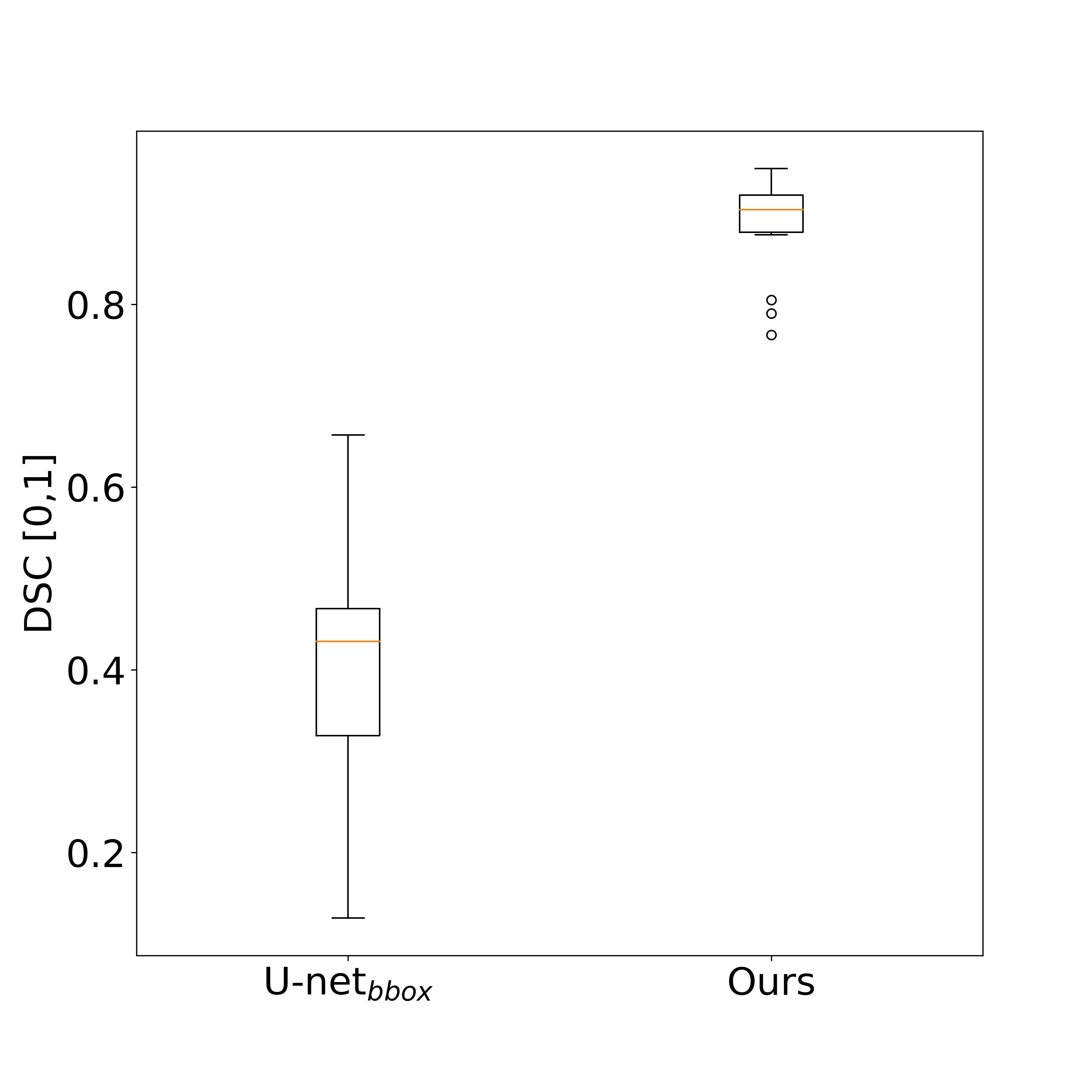}
			%\caption{fig2}
			\label{spleen_box_2}
		\end{minipage}
	}%
	\subfigure[]{
		\begin{minipage}[t]{0.18\linewidth}
			\centering
			\includegraphics[width=1.3 in]{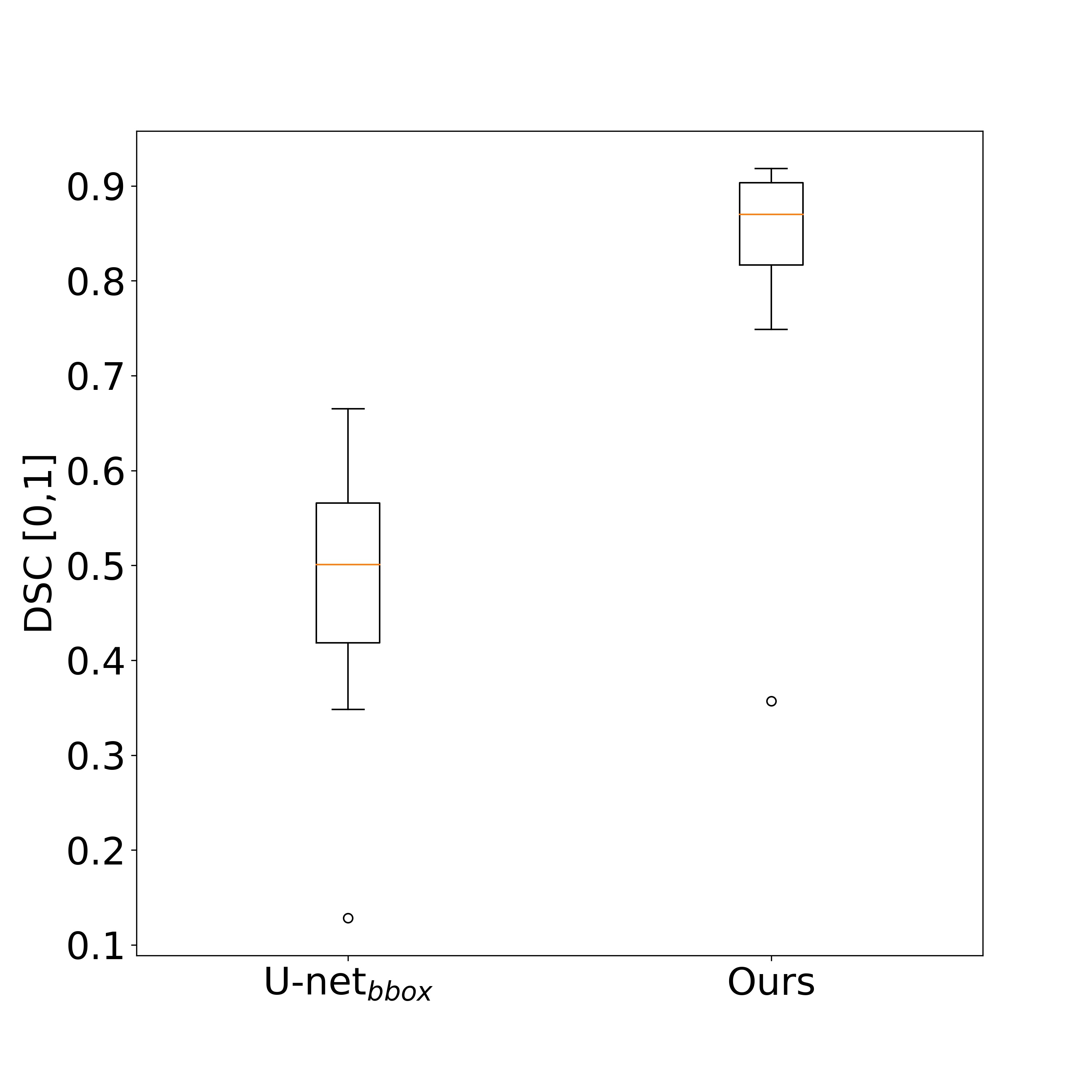}
			%\caption{fig2}
			\label{spleen_box_3}
		\end{minipage}
	}%
	\centering
	\caption{Comparative Dice distribution of different methods for all segmentation tasks. Row 1 is liver fold 1-5, row 2 is kidneys fold 1-5 and row 3 is spleen fold 1-3.}
	\label{box_img}
\end{figure}

Table.\ref{compare-kidneys} summaries the results on each fold and the averaged scores for kidneys. The final mean DSC of BA-Unet$_{bbox}$, GrabCut, BA-Unet$_{GC}$ and ours were 71.49$\pm$7.72, 76.23$\pm$16.88, 81.75$\pm$12.03 and 91.45$\pm$7.31, respectively, and the Jaccard scores were $56.16\pm 8.63$, 64.79$\pm$18.88, 70.73$\pm$14.99 and 85.01$\pm$10.93, respectively. the DSC distribution for all methods on each validation fold is presented in Fig. \ref{kits_box_1}--\ref{kits_box_5}. 

\begin{table}[!htp]
	\centering
	\setlength{\tabcolsep}{0.05mm}{
		\begin{tabular}{@{}lllll@{}}
			\toprule
			& BA-Unet$_{Bbox}  $                                                       & GrabCut & BA-Unet$_{GC}$ & Ours                                                                        \\ \midrule
			Fold 1 & \begin{tabular}[c]{@{}l@{}}67.14(11.53)/\\ 51.68(12.73)\end{tabular} & N/A     & N/A        & \begin{tabular}[c]{@{}l@{}}92.95(2.94)/\\ 86.97(5.04)\end{tabular}          \\
			Fold 2 & \begin{tabular}[c]{@{}l@{}}54.76(17.13)/\\ 39.54(15.66)\end{tabular} & N/A     & N/A        & \begin{tabular}[c]{@{}l@{}}93.88(3.23)/\\ 88.63(5.57)\end{tabular}          \\
			Fold 3 & \begin{tabular}[c]{@{}l@{}}63.54(13.85)/\\ 47.90(13.23)\end{tabular} & N/A     & N/A        & \begin{tabular}[c]{@{}l@{}}89.49(10.66)/\\ 82.31(13.90)\end{tabular}        \\ \midrule
			Avg    & \begin{tabular}[c]{@{}l@{}}61.83(14.17)/\\ 46.37(13.87)\end{tabular} & N/A     & N/A        & \textbf{\begin{tabular}[c]{@{}l@{}}92.11(5.61)/\\ 85.97(8.19)\end{tabular}} \\ \bottomrule
	\end{tabular}}
	\caption{Comparison of different methods on spleen segmentation. All measurements are reported as fold-averaged (standard variation) DSC and Jaccard scores [\%]. The best scores are marked in bold.}
	\label{compare-spleen}
\end{table}

Table.\ref{compare-spleen} shows the results on each fold and the averaged scores for spleens. The final mean DSC of BA-Unet$_{bbox}$ and ours were 61.83$\pm$14.17 and 92.11$\pm$5.61, and the Jaccard scores were 46.37$\pm$13.87 and 85.97$\pm$8.19, respectively. Fig. \ref{spleen_box_1}--\ref{spleen_box_3} depict the DSC distribution for all methods on each validation fold.

Table.\ref{compare-liver} illustrates the results on each fold and the averaged scores for livers. The final mean DSC of BA-Unet$_{bbox}$, GrabCut, BA-Unet$_{GC}$ and ours were 72.43$\pm$7.72, 84.08$\pm$6.80, $91.57\pm3.41$ and 95.19$\pm$0.84, respectively, and the Jaccard scores were 56.91$\pm$4.49, 73.14$\pm$9.74, 84.61$\pm$5.44 and 90.83$\pm$1.51, respectively. Fig. \ref{liver_box_1}--\ref{liver_box_5} show the DSC distribution for all methods on each validation fold.

\begin{table}[t]
	\centering
	\setlength{\tabcolsep}{0.05mm}{
		\begin{tabular}{@{}lllll@{}}
			\toprule
			& BA-Unet$_{Bbox}$                                                       & GrabCut                                                             & BA-Unet$_{GC}$                                                         & Ours                                                                        \\ \midrule
			Fold 1 & \begin{tabular}[c]{@{}l@{}}72.22(5.05)/\\ 56.76(6.05)\end{tabular} & \begin{tabular}[c]{@{}l@{}}85.95(6.51)/\\ 75.91(9.68)\end{tabular}  & \begin{tabular}[c]{@{}l@{}}90.50(6.48)/\\ 83.21(9.45)\end{tabular} & \begin{tabular}[c]{@{}l@{}}95.02(1.05)/\\ 90.52(1.88)\end{tabular}          \\
			Fold 2 & \begin{tabular}[c]{@{}l@{}}72.29(2.91)/\\ 56.68(3.54)\end{tabular} & \begin{tabular}[c]{@{}l@{}}83.71(6.55)/\\ 72.53(9.59)\end{tabular}  & \begin{tabular}[c]{@{}l@{}}91.46(2.51)/\\ 84.36(4.28)\end{tabular} & \begin{tabular}[c]{@{}l@{}}94.97(0.82)/\\ 90.44(1.48)\end{tabular}          \\
			Fold 3 & \begin{tabular}[c]{@{}l@{}}72.76(3.33)/\\ 57.30(4.10)\end{tabular} & \begin{tabular}[c]{@{}l@{}}85.23(6.09)/\\ 74.75(9.07)\end{tabular}  & \begin{tabular}[c]{@{}l@{}}92.57(2.59)/\\ 86.27(4.42)\end{tabular} & \begin{tabular}[c]{@{}l@{}}95.11(0.64)/\\ 90.68(1.77)\end{tabular}          \\
			Fold 4 & \begin{tabular}[c]{@{}l@{}}73.05(4.20)/\\ 57.71(5.12)\end{tabular} & \begin{tabular}[c]{@{}l@{}}83.88(7.50)/\\ 72.90(10.23)\end{tabular} & \begin{tabular}[c]{@{}l@{}}91.24(3.33)/\\ 84.06(5.44)\end{tabular} & \begin{tabular}[c]{@{}l@{}}95.45(1.09)/\\ 91.32(1.97)\end{tabular}          \\
			Fold 5 & \begin{tabular}[c]{@{}l@{}}71.81(3.05)/\\ 56.10(3.63)\end{tabular} & \begin{tabular}[c]{@{}l@{}}81.65(7.33)/\\ 69.62(10.14)\end{tabular} & \begin{tabular}[c]{@{}l@{}}91.94(2.16)/\\ 85.15(3.62)\end{tabular} & \begin{tabular}[c]{@{}l@{}}95.40(0.58)/\\ 91.21(1.06)\end{tabular}          \\ \midrule
			Avg    & \begin{tabular}[c]{@{}l@{}}72.43(3.71)/\\ 56.91(4.49)\end{tabular} & \begin{tabular}[c]{@{}l@{}}84.08(6.80)/\\ 73.14(9.74)\end{tabular}  & \begin{tabular}[c]{@{}l@{}}91.57(3.41)/\\ 84.61(5.44)\end{tabular} & \textbf{\begin{tabular}[c]{@{}l@{}}95.19(0.84)/\\ 90.83(1.51)\end{tabular}} \\ \bottomrule
	\end{tabular}}
	\caption{Comparison of different methods on liver segmentation. All measurements are reported as fold-averaged (standard variation) DSC and Jaccard scores [\%]. The best scores are marked in bold.}
	\label{compare-liver}
\end{table}

\section{Discussion}
\label{sec-6}

Accurate segmentation of CT images is essential in clinical diagnosis. In this paper, we propose an automatic segmentation method for CT images based on weak-supervised learning. Pixel-by-pixel segmentation results can be obtained using only a limited number of bounding box annotations. To demonstrate the effectiveness of the method, quantitative comparisons are performed on four datasets for three organs.

\subsection{Comparison with other methods}
Table.\ref{compare-kidneys}-\ref{compare-liver} show the comparative numerical results. Our method leads in both DSC and Jaccard scores for each selected organ. And, unlike other methods that suffer from severe inconsistency, the experimental results of our method have the smallest standard deviations except for the Jaccard score for kidneys. 

\begin{figure*}[h]
	\centering
	\subfigure[]{
		\begin{minipage}[t]{0.14\linewidth}
			\centering
			\includegraphics[width=0.9in]{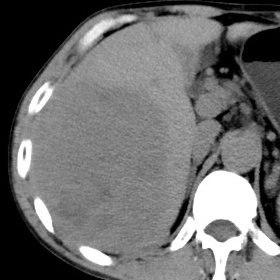}
			%\caption{fig1}
			\label{liver}
		\end{minipage}%
	}%
	\subfigure[]{
		\begin{minipage}[t]{0.13\linewidth}
			\centering
			\includegraphics[width=0.9in]{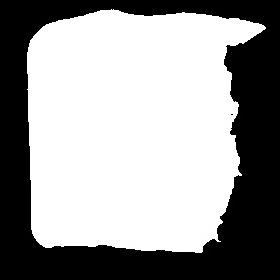}
			%\caption{fig2}
			\label{liver-bbox}
		\end{minipage}
	}%
	\subfigure[]{
		\begin{minipage}[t]{0.13\linewidth}
			\centering
			\includegraphics[width=0.9in]{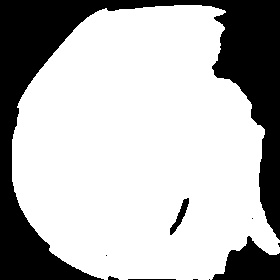}
			%\caption{fig2}
			\label{liver-graphcut}
		\end{minipage}
	}%
	\subfigure[]{
		\begin{minipage}[t]{0.13\linewidth}
			\centering
			\includegraphics[width=0.9in]{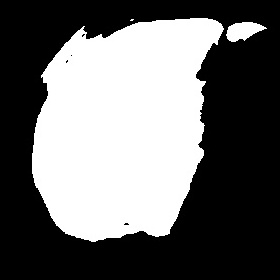}
			%\caption{fig2}
			\label{liver-gccnn}
		\end{minipage}
	}%
	\subfigure[]{
		\begin{minipage}[t]{0.14\linewidth}
			\centering
			\includegraphics[width=0.9in]{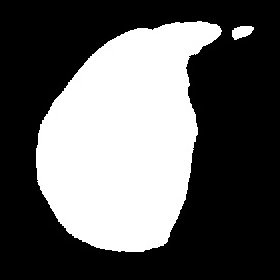}
			%\caption{fig2}
			\label{liver-result}
		\end{minipage}%
	}%
	\subfigure[]{
		\begin{minipage}[t]{0.13\linewidth}
			\centering
			\includegraphics[width=0.9in]{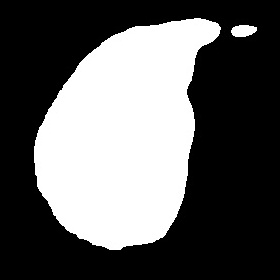}
			%\caption{fig2}
			\label{liver-gt}
		\end{minipage}
	}%
	%这个回车键很重要 \quad也可以
	
	\subfigure[]{
		\begin{minipage}[t]{0.13\linewidth}
			\centering
			\includegraphics[width=0.9 in]{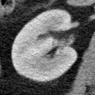}
			%\caption{fig2}
			\label{kits-left}
		\end{minipage}
	}%
	\subfigure[]{
		\begin{minipage}[t]{0.13\linewidth}
			\centering
			\includegraphics[width=0.9 in]{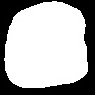}
			%\caption{fig2}
			\label{kits-bbox-left}
		\end{minipage}
	}%
	\subfigure[]{
		\begin{minipage}[t]{0.13\linewidth}
			\centering
			\includegraphics[width=0.9 in]{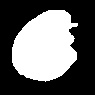}
			%\caption{fig2}
			\label{kits-graphcut-left}
		\end{minipage}
	}%
	\subfigure[]{
		\begin{minipage}[t]{0.13\linewidth}
			\centering
			\includegraphics[width=0.9 in]{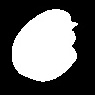}
			%\caption{fig2}
			\label{kits-gccnn-left}
		\end{minipage}
	}%
	\subfigure[]{
		\begin{minipage}[t]{0.13 \linewidth}
			\centering
			\includegraphics[width=0.9 in]{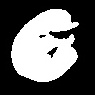}
			%\caption{fig2}
			\label{kits-result-left}
		\end{minipage}
	}%
	\subfigure[]{
		\begin{minipage}[t]{0.14\linewidth}
			\centering
			\includegraphics[width=0.9 in]{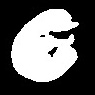}
			%\caption{fig2}
			\label{kits-gt-left}
		\end{minipage}
	}%
	
	\subfigure[]{
		\begin{minipage}[t]{0.13\linewidth}
			\centering
			\includegraphics[width=0.9 in]{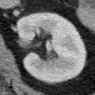}
			%\caption{fig2}
			\label{kits-right}
		\end{minipage}
	}%
	\subfigure[]{
		\begin{minipage}[t]{0.13\linewidth}
			\centering
			\includegraphics[width=0.9 in]{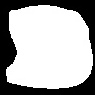}
			%\caption{fig2}
			\label{kits-bbox-right}
		\end{minipage}
	}%
	\subfigure[]{
		\begin{minipage}[t]{0.13\linewidth}
			\centering
			\includegraphics[width=0.9 in]{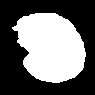}
			%\caption{fig2}
			\label{kits-graphcut-right}
		\end{minipage}
	}%
	\subfigure[]{
		\begin{minipage}[t]{0.13\linewidth}
			\centering
			\includegraphics[width=0.9 in]{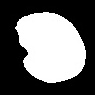}
			%\caption{fig2}
			\label{kits-gccnn-right}
		\end{minipage}
	}%
	\subfigure[]{
		\begin{minipage}[t]{0.13 \linewidth}
			\centering
			\includegraphics[width=0.9 in]{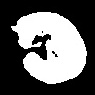}
			%\caption{fig2}
			\label{kits-result-right}
		\end{minipage}
	}%
	\subfigure[]{
		\begin{minipage}[t]{0.14\linewidth}
			\centering
			\includegraphics[width=0.9 in]{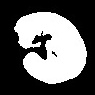}
			%\caption{fig2}
			\label{kits-gt-right}
		\end{minipage}
	}%
	
	\subfigure[]{
		\begin{minipage}[t]{0.14\linewidth}
			\centering
			\includegraphics[width=0.9 in]{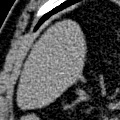}
			%\caption{fig1}
			\label{spleen}
		\end{minipage}%
	}%
	\subfigure[]{
		\begin{minipage}[t]{0.13\linewidth}
			\centering
			\includegraphics[width=0.9 in]{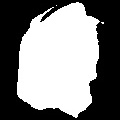}
			%\caption{fig2}
			\label{spleen-bbox}
		\end{minipage}
	}%
	\subfigure[]{
		\begin{minipage}[t]{0.14\linewidth}
			\centering
			\includegraphics[width=0.9 in]{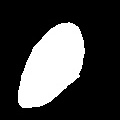}
			%\caption{fig2}
			\label{spleen-result}
		\end{minipage}%
	}%
	\subfigure[]{
		\begin{minipage}[t]{0.14\linewidth}
			\centering
			\includegraphics[width=0.9 in]{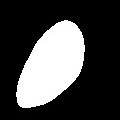}
			%\caption{fig2}
			\label{spleen-gt}
		\end{minipage}
	}%
	%这个回车键很重要 \quad也可以
	\centering
	\caption{Example segmentation results (only the target organs and their segmentation results are displayed): Row 1 (liver), row 2 (left kidney), row 3 (right kidney) from left to right: (1) Original CT slice, (2) BA-Unet$_{bbox}$, (3) GrabCut, (4) BA-Unet$_{GC}$ (5) Ours (6) Ground truth; Row 4 (spleen) from left to right: (1) Original CT scan, (2)BA-Unet$_{bbox}$, (3) Ours, (4) Ground truth.}
	\label{example.pic}
\end{figure*}

Fig. \ref{example.pic} shows some example segmentation results for all methods. We can easily obtain the differences in the results of the different methods. BA-Unet$_{bbox}$ can only tell the rough location of the target, and the output is more like object detection rather than semantic segmentation (Fig. \ref{liver-bbox}, Fig. \ref{kits-bbox-left}, Fig. \ref{kits-bbox-right} and Fig. \ref{spleen-bbox}). GrabCut, as described above, is very sensitive to the input image. When the boundary of the target to be divided is clear, it can depict the edges and shape with some accuracy (Fig. \ref{kits-graphcut-left}, Fig. \ref{kits-graphcut-right}). However, for liver CT slice (\ref{liver-graphcut}), the fuzzy boundaries of the organ lead to an excessive number of false positives. Whereas for spleen CT, GrabCut is useless for most samples. BA-Unet$_{GC}$, which is trained with the output of GrabCut as initial labels, can refine the segmentation results and makes the boundaries smoother. Nevertheless, it is struggled to extract the internal structures of organs, such as holes and depressions in the kidneys (Fig. \ref{liver-graphcut}, Fig. \ref{kits-gccnn-left} and Fig. \ref{kits-gccnn-right}). Compared to these methods, ours not only effectively depicts the outline of the organ but also well predicts the holes and sunken areas (Fig. \ref{liver-result}, Fig. \ref{kits-result-left}, Fig. \ref{kits-result-right} and Fig. \ref{spleen-result}).

\subsection{Comparison with related studies}

To further assess the performance of our proposed method, we compared it to another automatic segmentation method. To our best knowledge, few articles discuss weakly supervised learning for organ segmentation in CT images and use the same dataset as ours for testing. Thus, we can only ensure that the dataset used in the selected comparative article intersects as much as possible with the one we use, given that it discusses the same organs (liver, spleen and kidneys) under the same modality (CT).

We compared our method with the one proposed by Ma et al.\cite{ma2020abdomenct}, which presented an abdominal CT organ segmentation dataset including liver, kidney, spleen and pancreas, and provided benchmarks for fully supervised, semi-supervised and weakly supervised multi-organ segmentation models. They trained a 2D nn-UNet with MSD spleen dataset with annotations of sparse labels for weakly supervised learning. Under 50\% annotation rate, the model could obtain averaged DSC of 92.8 $\pm$5.48, 90.5$\pm$14.6 and 84.2$\pm$12.7 for liver, spleen and kidneys respectively on a mixture of 50 cases from LiTS and KiTS dataset. It can be seen from Table.\ref{related-wrok} that our method outperforms the weakly supervised learning benchmark proposed by Ma et al. in all three different segmentation tasks and leads by a large margin in kidney segmentation. 

\begin{table}[!htp]
	\centering
	
	\label{related-wrok}
	\begin{tabular}{@{}llll@{}}
		\toprule
		& Liver & Spleen & Kidneys \\ \midrule
		Ma et al.\cite{ma2020abdomenct}      & 92.8  & 90.5   & 84.2    \\
		Ours           & 95.19 & 92.11  & 91.45   \\ \bottomrule
	\end{tabular}
	\caption{Comparison with other automatic liver segmentation methods. All measeurements are reported as averaged DSC [\%]. }
\end{table}

\subsection{Differences in liver, spleen and kidneys segmentation performance} 
From our experimental results, we found that the accuracy of liver segmentation was slightly higher than that of kidneys and spleen. There are two potential reasons for this phenomenon. Firstly, the differences in organ morphology make the performance of feature extraction different. For example, the common shadow areas, holes, and depressed areas inside kidneys will increase the difficulty of feature extraction. Moreover, more high-dimensional information could be extracted from their images by a deep convolutional neural network than those of spleens and kidneys due to the larger volumes of livers. The second is the difference in data quality and quantity. The total number of cases in the MSD spleen dataset (41 cases) is much smaller than that in our liver (245 cases) and kidney (210 cases) datasets, and the organs in the images contain a lot of noise, which affects the final segmentation performance.

\subsection{Limitations}

Our method showed reliable results; however, it had some limitations, mainly about target boundaries predictions. First of all, as one can see from fig. \ref{example.pic}, for organs such as the livers, which have blurred boundaries with the surrounding tissues, there is still room for improvement in the shapes of these locations. Second, prediction accuracy is vulnerable to noises near the boundaries of organs, which are more inclined to be misclassified as positive. These drawbacks that affect the segmentation accuracy will be our further work in the future.

\section{Conclusions}
\label{sec-7}
Deep learning methods have been demonstrated to achieve excellent accuracy in CT image segmentation, but they also bring huge demand for accurate annotation. In this study, we have proposed a fully automated CT segmentation approach based on weakly supervised learning. It performs k-means clustering of weak labels in the form of bounding boxes to generate pseudo-mask and iteratively trains the convolutional neural network, which can produce out precise segmentation predictions. We have trained and validated on four datasets containing three organs with 627 CT images, and the results show that the method achieves a high level of accuracy and preferable robustness. It is attractive to validate our method on more clinical data, and different segmentation works in the future.

\section*{Acknowledgment}
The authors thank the organizers of the LiTS challenge, KiTS19 challenge, and MSD segmentation challenge for providing CT segmentation data. This work was supported in part by the National Natural Science Foundation of China (Grant No. 12090020 and No. 12090025), and in part by Zhejiang Provincial Natural Science Foundation of China under Grant No. LSD19H180005.

\bibliographystyle{elsarticle-num}
%\bibliography{./example.bib}


\end{document}